\documentclass[journal]{IEEEtran}
\ifCLASSINFOpdf
\else
\fi

\usepackage{graphicx}
\usepackage{url}
\usepackage{breakurl}
\usepackage{multirow}
\usepackage{array}
\usepackage{tabu}
\usepackage{xcolor}

\usepackage[hang,flushmargin]{footmisc}

\newcommand*\rot{\rotatebox{90}}

\makeatletter
\def\footnoterule{\relax%
 \kern-5pt
 \hbox to \columnwidth{\hfill\vrule width \columnwidth height 0.6pt\hfill}
 \kern4.6pt}
\makeatother

\begin{document}
\title{Fingerprint Spoof Buster}%:\\Liveness Detection using Local Patches}

\author{Tarang Chugh*,~\IEEEmembership{Student Member,~IEEE,}
        Kai Cao, and~Anil K. Jain,~\IEEEmembership{Life~Fellow,~IEEE}%
        
\thanks{T. Chugh, K. Cao, and A. K. Jain are with the Department of Computer Science and Engineering, Michigan State University, East Lansing, MI, 48824. E-mail: \{chughtar, kaicao, jain\}@cse.msu.edu}% <-this % stops a space
\thanks{*Corresponding Author}
\thanks{A preliminary version of this paper was presented at the International Joint Conference on Biometrics (IJCB), Denver, CO, October 1-4, 2017~\cite{chugh2017fingerprint}.}
}
\maketitle

\begin{abstract}
The primary purpose of a fingerprint recognition system is to ensure a reliable and accurate user authentication, but the security of the recognition system itself can be jeopardized by spoof attacks. This study addresses the problem of developing accurate, generalizable, and efficient algorithms for detecting fingerprint spoof attacks. Specifically, we propose a deep convolutional neural network based approach utilizing local patches centered and aligned using fingerprint minutiae. Experimental results on three public-domain LivDet datasets (2011, 2013, and 2015) show that the proposed approach provides state-of-the-art accuracies in fingerprint spoof detection for intra-sensor, cross-material, cross-sensor, as well as cross-dataset testing scenarios. For example, in LivDet 2015, the proposed approach achieves 99.03\% average accuracy over all sensors compared to 95.51\% achieved by the LivDet 2015 competition winners. Additionally, two new fingerprint presentation attack datasets containing more than 20,000 images, using two different fingerprint readers, and over 12 different spoof fabrication materials are collected. We also present a graphical user interface, called Fingerprint Spoof Buster, that allows the operator to visually examine the local regions of the fingerprint highlighted as live or spoof, instead of relying on only a single score as output by the traditional approaches.
\end{abstract}


% \fi
\IEEEpeerreviewmaketitle

\section{Introduction}
\label{sec:introduction}
\IEEEPARstart{W}{ITH} the ubiquitous deployment of fingerprint recognition systems in many day-to-day applications, such as financial transactions, international border security, unlocking a smartphone, etc., the vulnerability of the system security to \textit{presentation attacks} is of growing concern~\cite{marcel2014handbook,IARPAProject}. The ISO standard \textit{IEC 30107-1:2016(E)}~\cite{isopad} defines presentation attacks as the \textit{``presentation to the biometric data capture subsystem with the goal of interfering with the operation of the biometric system"}. These attacks can be realized through a number of methods including, but not limited to, use of (i)~\textit{gummy fingers}~\cite{matsumoto2002impact}, \textit{i.e.} fabricated finger-like objects with accurate imitation of another individual's fingerprint ridge-valley structures, (ii) \textit{2D or 3D printed fingerprint targets}~\cite{cao2016hacking, arora2016design, arora2017goldfingers}, (iii)~\textit{altered fingerprints}~\cite{yoon2012altered}, \textit{i.e.} intentionally tampered or damaged real fingerprint patterns to avoid identification, and (iv)~\textit{cadaver fingers}~\cite{marasco2015survey}. Among these, fingerprint spoof attacks (\textit{i.e.} gummy fingers and printed targets) are the most common form of presentation attacks, with a multitude of fabrication processes ranging from basic \textit{molding and casting} to utilizing sophisticated 2D and 3D printing techniques~\cite{matsumoto2002impact, cao2016hacking, arora2016design, arora2017goldfingers}.

\begin{figure}[t]
\centering
\includegraphics[width=\linewidth]{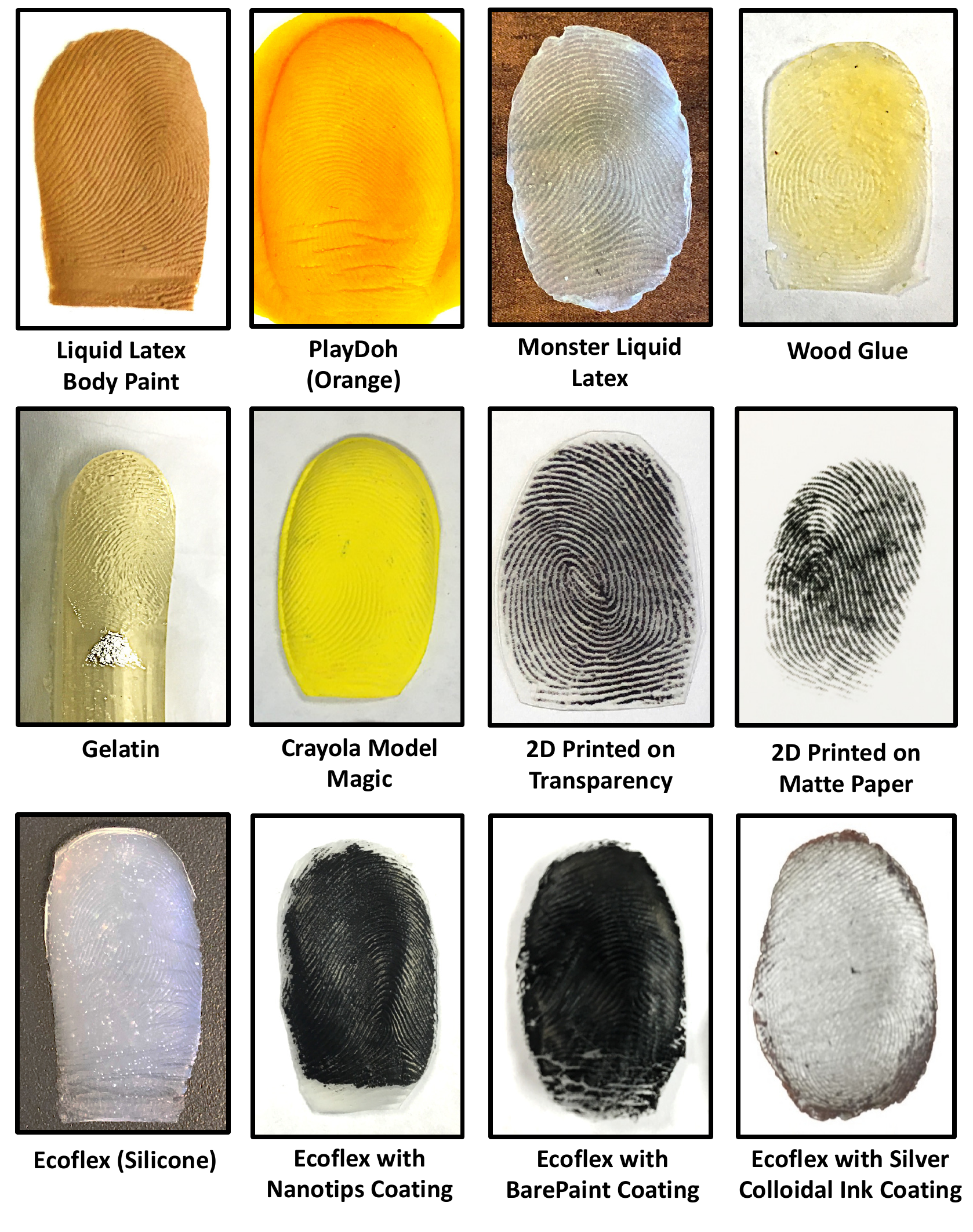}
\caption{Fingerprint spoof attacks can be realized using various readily available fabrication materials, such as PlayDoh, WoodGlue, Gelatin, etc.}
\label{fig:spoofs}
\end{figure}

\begin{figure*}[t]
\centering
\includegraphics[width=\linewidth]{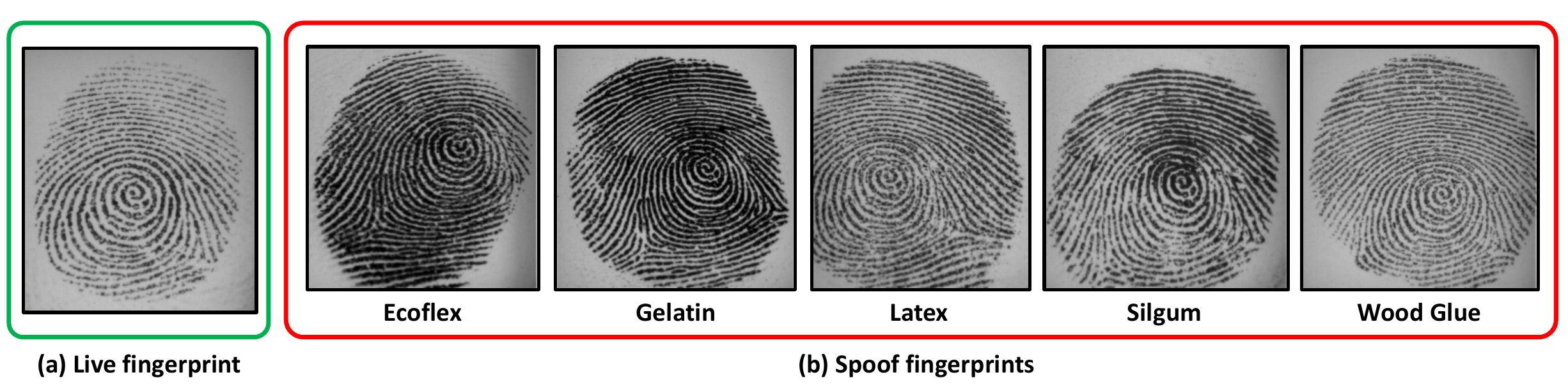}
\caption{Visual comparison between (a) a live Fingerprint, and (b) the corresponding spoofs (of the same live finger) made with different materials. Images are taken from LivDet-2011 dataset (Biometrika sensor)~\cite{yambay2012livdet}. Our method can successfully distinguish between live and spoof fingerprints. The spoofness scores for live fingerprint is $0.00$, and for spoof fingerprints are $0.95$, $0.97$, $0.99$, $0.99$, and $0.95$ for Ecoflex, Gelatin, Latex, Silgum, and Wood Glue, respectively.}
\label{fig:SpoofFpt}
\end{figure*}

\begin{table*}[t]
\caption{Performance comparison (Average Classification Error [\%]) of software-based spoof detection studies~\cite{nogueira2016fingerprint,gragnaniello2015investigation} on LivDet 2011, 2013, and 2015 datasets.}
\label{tab:performanceLivDet}
\centering
\resizebox{\linewidth}{!}{
\begin{tabular}{ | p{2.5 cm} | >{\centering\arraybackslash}p{6.4 cm} | >{\centering\arraybackslash}p{1.7cm} | >{\centering\arraybackslash}p{1.7cm} | >{\centering\arraybackslash}p{1.7cm} |}
\hline
\textbf{Study} & \textbf{Approach} & \textbf{LivDet 2011} & \textbf{LivDet 2013*} & \textbf{LivDet 2015} \\ \hline

Ghiani et al.~\cite{ghiani2012fingerprint} & Local Phase Quantization (LPQ) &  11.1 & 3.0 & N/A \\ \hline

Gragniello et al.~\cite{gragnaniello2013fingerprint} & Weber Local Descriptor (WLD) & 7.9  & N/A & N/A \\ \hline

Ghiani et al.~\cite{ghiani2013fingerprint} & Binarized Statistical Image Features (BSIF) & 7.2 & 2.1 & N/A \\ \hline

Gragniello et al.~\cite{gragnaniello2015local} & Local Contrast-Phase Descriptor (LCPD) & 5.7 & 1.3 & N/A \\ \hline

Nogueira et al.~\cite{nogueira2016fingerprint} & Transfer Learning + CNN-VGG + Whole Image & 4.5 & 1.1 & 4.5 \\ \hline

Pala et al.~\cite{pala2017deep} & Custom CNN with triplet loss + Randomly selected local patches & 3.33 & 0.58 & N/A \\ \hline

\textbf{Proposed Approach} & CNN-MobileNet-v1 + Minutiae-based local patches & \textbf{1.67} & \textbf{0.25} & \textbf{0.97} \\ \hline
\end{tabular}
}
\begin{flushleft}
{\footnotesize *LivDet 2013 includes results for Biometrika and Italdata sensors.}
\end{flushleft}
\end{table*}

Commonly available materials, such as gelatin, silicone, play-doh, etc., have been utilized to generate fingerprint spoofs (see Figs.~\ref{fig:spoofs} and~\ref{fig:SpoofFpt}), capable of circumventing a fingerprint recognition system security with a reported success rate of more than $70\%$~\cite{biggio2012security}. For instance, in July 2016, researchers at Michigan State University unlocked a fingerprint secure-smartphone using a 2D printed fingerprint spoof to help police with a homicide case\footnote{http://statenews.com/article/2016/08/how-msu-researchers-unlocked-a-fingerprint-secure-smartphone-to-help-police-with-homicide-case}, using a technique proposed in~\cite{cao2016hacking}. In Sept. 2013, shortly after Apple released iPhone 5s with inbuilt TouchID fingerprint technology, Germany's Chaos Computer Club\footnote{http://www.ccc.de/en/updates/2013/ccc-breaks-apple-touchid} hacked its capacitive sensor by utilizing a high resolution photograph of the enrolled user's fingerprint to fabricate a spoof fingerprint with wood glue. In another incident, in March 2013, a Brazilian doctor was arrested for using spoof fingers made of silicone to fool the biometric attendance system at a hospital in Sao Paulo\footnote{\url{http://www.bbc.com/news/world-latin-america-21756709}}. Since then, many successful spoof attacks have highlighted the vulnerabilities of fingerprint biometric systems\footnote{http://fortune.com/2016/04/07/guy-unlocked-iphone-play-doh/}$^{,}$\footnote{https://srlabs.de/bites/spoofing-fingerprints/}.

%Cao and Jain~\cite{cao2016hacking} demonstrated a simple hack to bypass the biometric security of two smartphones by scanning and printing fingerprints using silver conductive ink on a special AgIC (silver ink circuit) paper.

%With the introduction of biometric based mobile payment application  In Sept. 2013, shortly after Apple released iPhone 5s with inbuilt TouchID fingerprint technology, Germany's Chaos Computer Club\footnote{http://www.ccc.de/en/updates/2013/ccc-breaks-apple-touchid} hacked its capacitive sensor by utilizing a high resolution photograph of the enrolled user's fingerprint to fabricate a spoof fingerprint with wood glue.  The mobile payment industry\footnote{http://www.biometricupdate.com/wp-content/uploads/2017/03/special-report-mobile-biometric-applications.pdf} is expected to grow from US\$250 billion in 2014, to US\$750 billion in annual transactions by 2020.

%\begin{figure}[t]
%\centering
%\includegraphics[width=\linewidth]{img/SpoofFpt4.pdf}
%\caption{Fingerprint presentation attacks can be realized using (a)~gummy fingers~\cite{matsumoto2002impact,yambay2012livdet}, (b) 2D or 3D printed fingerprint  targets~\cite{cao2016hacking, arora2016design, arora2017goldfingers}, (c)~altered fingers~\cite{yoon2012altered}, or (d)~cadaver fingers~\cite{marasco2015survey}.}
%\label{fig:pa}
%\end{figure}

\begin{figure}[htbp!]
\centering
\includegraphics[trim=0cm 0cm 0cm 0cm, width=0.92\linewidth]{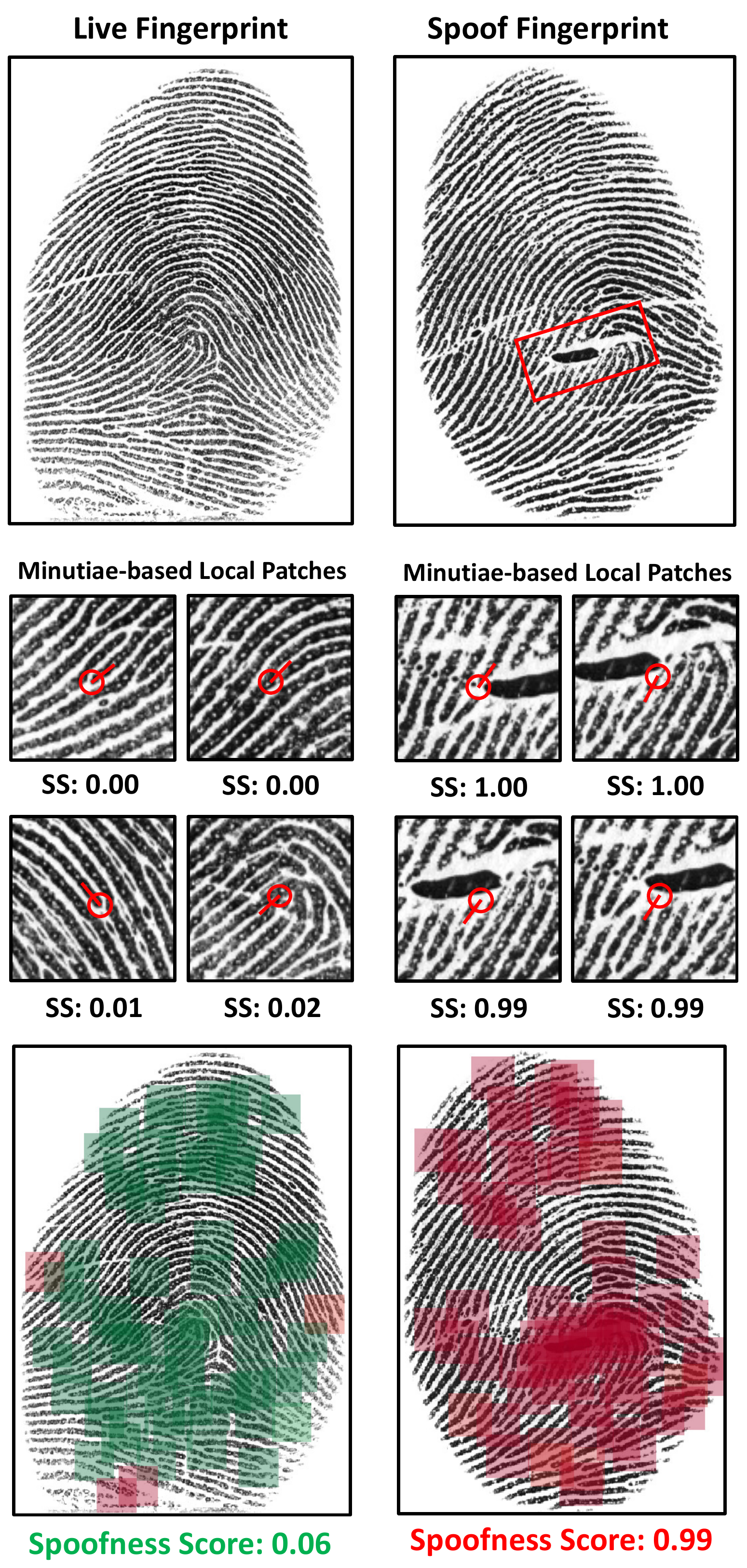}
\caption{Example of a live fingerprint and corresponding spoof fingerprint, with the artifacts introduced in the spoofs highlighted in red. The minutiae based local patches extracted around the artifacts are also presented. The images are taken from MSU Fingerprint Presentation Attack Dataset (MSU-FPAD) - CrossMatch Sensor and the spoof material used is Silicone (Ecoflex). The spoofness scores output by the proposed approach for the live and spoof fingerprints are 0.06 and 0.99, respectively.}
\label{fig:artifacts}
\vspace{-3mm}
\end{figure}

Fingerprint spoof detection methods are urgently needed to thwart such attacks on fingerprint authentication systems, thereby increasing the security and user confidence in such systems. The various anti-spoofing approaches proposed in the literature can be broadly classified into hardware-based and software-based solutions \cite{marcel2014handbook, marasco2015survey, mura2015livdet}. The hardware-based solutions typically require the fingerprint reader to be augmented with sensor(s) to detect the characteristics of vitality, such as blood flow~\cite{lapsley1998anti}, skin distortion~\cite{antonelli2006fake}, odor~\cite{baldisserra2006fake} and so on. There are also special types of fingerprint sensors, such as Lumidigm's multispectral scanner~\cite{rowe2006multispectral} and Compact Imaging's multiple reference optical coherence tomography (OCT)~\cite{hogan2015multiple}, that capture sub-dermal ridge patterns in the finger. An open-source fingerprint reader with a two-camera design provides two complementary streams of information useful for spoof detection~\cite{engelsma2017raspireader}. Software-based solutions, on the other hand, extract features from the presented fingerprint image (or a sequence of frames) acquired by the fingerprint sensors, without incurring any additional hardware cost, to differentiate between live and spoof fingers.

The software-based solutions published in the literature typically utilize one of the following approaches: (i)~anatomical features (e.g. pore locations and their distribution~\cite{marcialis2010analysis}), (ii)~physiological features (e.g. perspiration~\cite{marasco2012combining}), or (iii)~texture-based features (e.g. Local Phase Quantization (LPQ)~\cite{ghiani2012fingerprint}, Binarized Statistical Image Features (BSIF)~\cite{ghiani2013fingerprint}, and Weber Local Descriptor~\cite{gragnaniello2013fingerprint}). %Galbally et al.~\cite{galbally2012high} proposed a set of ten quality based features related to ridge strength, continuity, clarity, and integrity of the ridge-valley structure for spoof detection. 
Gragniello et al.~\cite{gragnaniello2015local} proposed a 2D local contrast-phase descriptor (LCPD), utilizing both spatial and frequency domain information. In contrast to the custom-tailored anti-spoof features, Menotti et al.~\cite{menotti2015deep}, Nogueira et al.~\cite{nogueira2016fingerprint}, and Pala et al.~\cite{pala2017deep} have proposed convolutional neural network (CNN) based solutions whose performances were shown to surpass many published spoof detection algorithms (see Table~\ref{tab:performanceLivDet}).

%%Spoof Detection as Open-set
%The performance of learning-based approaches, such as CNN, can be significantly affected by the fabrication materials used to generate spoofs.
One of the limitations of many of the published anti-spoof methods is their poor generalization performance across spoof materials. Studies in~\cite{nogueira2016fingerprint, marasco2011robustness, rattani2015open} have shown that when a spoof detector is evaluated on spoofs fabricated using materials that were not seen during training, there can be up to a three-fold increase in the spoof detection error rates. To generalize an algorithm's effectiveness across spoof fabrication materials, called \textit{cross-material} performance, some studies have approached spoof detection as an \textit{open-set problem}\footnote{Open-set problems address the possibility of spoof classes during testing, that were not seen during training. Closed-set problems, on the other hand, evaluate only those spoof classes that the system was trained on.}. Rattani et al.~\cite{rattani2015open} applied the Weibull-calibrated SVM (W-SVM), a variant of SVM based on properties of statistical extreme value theory, to detect spoofs made of new materials. Ding and Ross~\cite{ding2016ensemble} trained an ensemble of multiple one-class SVMs using textural features extracted from only live fingerprint images.

A series of fingerprint Liveness Detection (LivDet) competitions have been held since 2009 to advance state-of-the-art and benchmark the proposed anti-spoofing solutions~\cite{ghiani2017review}. The best performer in the most recent\footnote{Results of LivDet 2017 are expected to be presented at the $11^{th}$ IAPR International Conference on Biometrics (ICB), Brisbane, Feb. 20-23, 2018.} LivDet 2015~\cite{mura2015livdet}, Nogueira et al.~\cite{nogueira2016fingerprint}, utilized transfer learning, where deep CNNs originally designed for object recognition and pre-trained on ImageNet database~\cite{russakovsky2015imagenet}, were fine-tuned on fingerprint images to differentiate between live and spoof fingerprints. In their approach, the networks were trained on whole fingerprint images resized to $227 \times 227$ pixels for VGG~\cite{simonyan2014very} and $224 \times 224$ pixels for AlexNet~\cite{krizhevsky2012imagenet} as required by these networks. However, there are three disadvantages of using this approach: (i)~fingerprint images from some of the sensors used in LivDet datasets, such as Crossmatch L Scan Guardian ($640 \times 480$), have a large blank area ($\ge 50\%$) surrounding the friction ridge region. Directly resizing these images, from $640 \times 480$ to $227 \times 227$, eventually results in the friction ridge area occupying less than 10\% of the original image size; (ii)~resizing a rectangular image of size, say $w \times h$, to a square image, say $p \times p$, leads to different amounts of information retained in the two spatial image dimensions; (iii)~downsizing an image, in general, leads to a significant loss of discriminatory information.

It is important to consider various sources of noise involved in the spoof fabrication process itself that can introduce some artifacts, such as missing friction ridge regions, cracks, air bubbles, etc., in the spoofs. The primary consequence of such artifacts is the creation of spurious minutiae in the fingerprint images sensed from spoofs. The local regions around these spurious minutiae can, therefore, provide salient cues to differentiate a spoof fingerprint from live fingerprints (see Fig.~\ref{fig:artifacts}). We utilize this observation to train a two-class CNN using local patches around the extracted minutiae, as opposed to the whole fingerprint images or randomly selected local patches, to design a fingerprint spoof detector. We show that this approach is more robust to novel fabrication materials than earlier approaches that utilize the whole image~\cite{nogueira2016fingerprint} or randomly selected local patches~\cite{pala2017deep}. 

The proposed approach, utilizing local patches of size $p \times p$, ($p=96$), centered at minutiae, (i)~addresses the previously mentioned drawbacks of downsizing whole fingerprint images to train the CNN, (ii)~provides large amount of data (an average of 48 patches/fingerprint image), sufficient to train the deep CNN architectures from scratch without overfitting, (iii)~learns salient textural information from local regions, robust to differentiate between spoof and live fingerprints, and (iv)~provides a fine-grained representation of the fingerprint images capable of localizing partial spoof fingerprints and fingerprint alterations. The output of the CNN is a confidence score in the range [$0-1$], defined as \textit{Spoofness Score}; higher the spoofness score, more likely the patch is extracted from a spoof fingerprint. For a given image, the spoofness scores corresponding to the minutiae-based local patches are averaged to give the global spoofness score for the whole image. Furthermore, a fusion of CNN models trained on multi-scale patches (ranging in size from from $64 \times 64$ to $128 \times 128$), centered and aligned using minutiae, is shown to further boost the spoof detection performance.\\

\begin{figure*}[t]
\centering
\includegraphics[width=\linewidth]{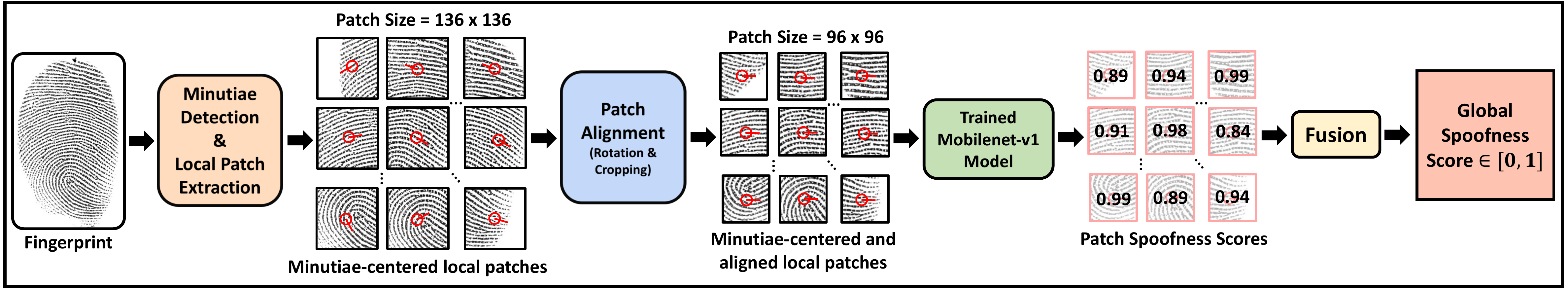}
\caption{An overview of the proposed approach for fingerprint spoof detection using CNNs trained on local patches centered and aligned using minutiae location and orientation, respectively.} %A typical live fingerprint image captured by CrossMatch Guardian 200 reader contains an average of 40 minutiae points. The number of minutiae points for the spoof images, also captured by CrossMatch Guardian 200 reader ranges between [??, ??].}
\label{fig:overview}
\end{figure*}

\begin{table*}[t]
\centering
\caption{Summary of the Liveness Detection (LivDet) datasets utilized in this study.}
\label{tab:datasets}
\resizebox{\linewidth}{!}{
\begin{tabular}{ | p{2 cm} || >{\centering\arraybackslash}p{1.3 cm} | >{\centering\arraybackslash}p{1.3 cm} | >{\centering\arraybackslash}p{1.3 cm} | >{\centering\arraybackslash}p{1.3 cm} || >{\centering\arraybackslash}p{1.3 cm} | >{\centering\arraybackslash}p{1.3 cm} %| >{\centering\arraybackslash}p{2.4 cm} 
|| >{\centering\arraybackslash}p{1.5 cm} | >{\centering\arraybackslash}p{1.5 cm} | >{\centering\arraybackslash}p{1.6 cm} | >{\centering\arraybackslash}p{2.0 cm} |}
\hline
\textbf{Dataset} & \multicolumn{4}{c||}{\textbf{LivDet 2011~\cite{yambay2012livdet}}} & \multicolumn{2}{c ||}{\textbf{LivDet 2013~\cite{ghiani2013livdet}}} & \multicolumn{4}{c|}{\textbf{LivDet 2015~\cite{mura2015livdet}}} \\ \hline

\textbf{Fingerprint Reader} & Biometrika & ItalData & Digital Persona &  Sagem  & Biometrika & ItalData %& Crossmatch 
& GreenBit &  Biometrika & Digital Persona & CrossMatch   \\ \hline

\textbf{Model} & FX2000 & ET10 & 4000B &  MSO300 & FX2000 & ET10 %& L Scan Guardian 
& DactyScan26 &  HiScan-PRO & U.are.U 5160 & L Scan Guardian  \\ \hline

\textbf{Image Size} & $315 \times 372$ & $640 \times 480$ & $355 \times 391$ &  $352 \times 384$ & $ 315 \times 372 $ & $640 \times 480$ %& $800 \times 750$ 
& $500 \times 500$  &  $1000 \times 1000$ & $252 \times 324$ & $640 \times 480$  \\ \hline

\textbf{Resolution (dpi)} & $500$ & $500$ & $500$ & $500$ & $569$ & $500$ %& $500$ 
& $500$ & $1000$ & $500$ & $500$  \\ \hline

\textbf{\#Live Images} & \multirow{2}{*}{1000/1000} & \multirow{2}{*}{$1000 / 1000$} & \multirow{2}{*}{$1000 / 1000$} & \multirow{2}{*}{$1000 / 1000$} & \multirow{2}{*}{$1000 / 1000$} & \multirow{2}{*}{$1000/1000$} %& \multirow{2}{*}{$1250/1250$}  
& \multirow{2}{*}{$1000 / 1000$} &  \multirow{2}{*}{$1000/1000$} & \multirow{2}{*}{$1000 / 1000$} & \multirow{2}{*}{$1510 / 1500$}    \\ 
\textbf{ Train / Test } & & & & & & & & & & \\ \hline

\textbf{\#Spoof Images} & \multirow{2}{*}{1000/1000} & \multirow{2}{*}{$1000 / 1000$} & \multirow{2}{*}{$1000 / 1000$} & \multirow{2}{*}{$1000 / 1000$} & \multirow{2}{*}{$1000 / 1000$} &\multirow{2}{*}{$1000 / 1000$} %& \multirow{2}{*}{1000/1000} 
& \multirow{2}{*}{$1000 / 1500$} &  \multirow{2}{*}{$1000 / 1500$} & \multirow{2}{*}{$1000 / 1500$} & \multirow{2}{*}{$1473 / 1448$}   \\  
\textbf{ Train / Test } & & & & & & & & & & \\ \hline

\textbf{Cooperative Subject*} & Yes & Yes & Yes & Yes & No & No & Yes & Yes & Yes & Yes  \\ \hline

\textbf{Spoof Materials} & \multicolumn{2}{>{\centering\arraybackslash}p{2.6cm}|}{Ecoflex, Gelatine, Latex, Silgum, Wood Glue}  & \multicolumn{2}{>{\centering\arraybackslash}p{2.7cm}||}{Gelatine, Latex, Play Doh, Silicone, Wood Glue} & \multicolumn{2}{>{\centering\arraybackslash}p{2.7cm}||}{Ecoflex, Gelatine, Latex, Modasil, Wood Glue} %& Body Double, Latex, Play Doh, Wood Glue 
& \multicolumn{3}{>{\centering\arraybackslash}p{5.1cm}|}{Ecoflex, Gelatine, Latex, Wood Glue, Liquid Ecoflex, RTV} &  Body Double, Ecoflex, Play Doh, OOMOO, Gelatin\\ \hline
\end{tabular}
}
%\vspace{-2mm}
\begin{flushleft}
{\footnotesize *In the cooperative subject scenario, a subject willingly provides a negative impression of the fingerprint as a mold, while in the non-cooperative subject scenario, the fingerprint mold is created by using the latent fingerprint lifted off a surface touched by the subject.\\LivDet datasets are available to interested researchers at~\url{http://livdet.org/registration.php}. }
\end{flushleft}
%\vspace{-2mm}
\end{table*}

\begin{table*}[t]
\centering
\caption{Summary of the MSU Fingerprint Presentation Attack Dataset (MSU-FPAD) and Precise Biometrics Spoof-Kit Dataset (PBSKD).}
\label{tab:datasetsMSU}
\resizebox{\linewidth}{!}{
\begin{tabular}{ | p{3 cm} || >{\centering\arraybackslash}p{3cm} | >{\centering\arraybackslash}p{3cm} || >{\centering\arraybackslash}p{5.1cm} | >{\centering\arraybackslash}p{5.1cm} |}
\hline
\textbf{Dataset} & \multicolumn{2}{c||}{\textbf{MSU-FPAD}} & \multicolumn{2}{c|}{\textbf{Precise Biometrics Spoof-Kit}}\\ \hline

\textbf{Fingerprint Reader} & CrossMatch & Lumidigm & CrossMatch & Lumidigm \\ \hline

\textbf{Model} & Guardian 200 & Venus 302 & Guardian 200 & Venus 302 \\ \hline

\textbf{Image Size} & $750 \times 800$ & $400 \times 272$ & $750 \times 800$ & $400 \times 272$ \\ \hline

\textbf{Resolution (dpi)} & $500$ & $500$ & $500$ & $500$ \\ \hline

\textbf{\#Live Images} & \multirow{2}{*}{$2,250$ / $2,250$} & \multirow{2}{*}{$2,250$ / $2,250$} & \multirow{2}{*}{$250$ / $250^{\dagger}$} & \multirow{2}{*}{$250$ / $250^{\dagger}$} \\
\textbf{ Train / Test } & & & & \\ \hline

\textbf{\#Spoof Images} & \multirow{2}{*}{$3,000$ / $3,000$} & \multirow{2}{*}{$2,250$ / $2,250$} & \multirow{2}{*}{$250$ / $250$} & \multirow{2}{*}{$200$ / $200^\ddagger$} \\
\textbf{ Train / Test } & & & &\\ \hline

\textbf{Cooperative*} & Yes & Yes & Yes & Yes \\ \hline

\textbf{Spoof Materials} & \multicolumn{2}{>{\centering\arraybackslash}p{6cm}||}{Ecoflex, PlayDoh, 2D Print (Matte Paper), 2D Print (Transparency)} & \multicolumn{2}{>{\centering\arraybackslash}p{10.2cm}|}{Ecoflex, Gelatin, Latex body paint, Ecoflex with silver colloidal ink coating, Ecoflex with BarePaint coating, Ecoflex with Nanotips coating, Crayola Model Magic, Wood glue, Monster Liquid Latex, and 2D printed fingerprint on office paper} \\ \hline
\end{tabular}
}
\begin{flushleft}
{\footnotesize $\dagger$1000 randomly sampled live fingerprint images from MSU-FPAD are selected for Precise Biometrics Spoof-Kit Dataset.\\
$\ddagger$ Lumidigm fingerprint reader does not image Silicone (EcoFlex) spoofs with NanoTips and BarePaint coatings.}
\end{flushleft}
\end{table*}

The main contributions of this study are enumerated below. They also highlight the differences between this paper and our preliminary study~\cite{chugh2017fingerprint}:
\begin{itemize}
\item Utilized fingerprint domain-knowledge to design a robust fingerprint spoof detector, where local patches centered and aligned using fingerprint minutiae are utilized for training a CNN model. This differs from other published approaches which have generally used the whole fingerprint image for spoof detection.%, to achieve a 55\% reduction in intra-sensor, a 78\% reduction in cross-material, and a 31\% reduction in cross-dataset state-of-the-art error rates.
\item A fine-grained representation of fingerprint images using local patches %, compared to existing global image-based approaches, 
capable of detecting partial spoof fingerprint regions and fingerprint alterations.
 \item Experimental results on publicly available datasets (LivDet 2011, 2013, and 2015), including intra-sensor, cross-material, cross-sensor, and cross-dataset scenarios, show that the proposed approach outperforms the state-of-the-art results published on these three datasets. For example, in LivDet 2015, our algorithm achieves 99.03\% average accuracy over all sensors compared to 95.51\% achieved by the LivDet 2015 winner~\cite{mura2015livdet}.
%\end{itemize}
%The major differences between this paper and our preliminary study published in~\cite{chugh2017fingerprint} are as follows:
%\begin{itemize}
%\item Improved state-of-the-art spoof detection results on LivDet datasets by aligning local patches based on minutiae orientation.
\item Real-time spoof detection by using MobileNet-v1~\cite{howard2017mobilenets}, compared to Inception-v3~\cite{szegedy2016rethinking} used in the preliminary study~\cite{chugh2017fingerprint}. The average classification time for a single input image on a NVIDIA GTX 1080 Ti GPU is 100ms compared to 800ms for Inception-v3 network.
 %MobileNet-v1 is a low-latency network utilizing depthwise separable convolutions~\cite{chollet2016xception} containing 4.2 million parameters, compared to 23.2 million parameters in Inception-v3. 
% to 100ms for MobileNet-v1 network.
\item Collected two new fingerprint presentation attack datasets\footnote{These datasets will be made available to interested researchers once this paper is accepted for publication.} containing more than $20,000$ fingerprint images, using two different fingerprint readers and over 12 different spoof fabrication materials.
%A large dataset, called the MSU Fingerprint Presentation Attack Dataset (MSU FPAD), including a total of 9,000 live and 10,500 spoof fingerprint images, collected using 2 different optical sensors (CrossMatch Guardian 200 and Lumidigm Venus 302) from 50 subjects and using 4 spoof materials (Silicone, PlayDoh, 2D printed on matte paper, 2D printed on transparency).
 \item Developed a graphical user interface (GUI) for real-time fingerprint spoof detection, called Fingerprint Spoof Buster, which allows the fingerprint reader operator to visually examine the local regions of the fingerprint highlighted as live or spoof, instead of relying on a single score, as output by the traditional approaches.
 %The average classification time for a single input image on a Nvidia GTX 1080Ti GPU is 100ms and on an Intel i7 (3.3 GHz) processor is 1s.
% \item The MobileNet-v1 network was modified to include center loss~\cite{wen2016discriminative} to increase the inter-class separation and make it robust against  challenging spoofs. The modified network is also observed to be robust against unknown spoof types.
%\item
\end{itemize}

%\textcolor{red}{Roy et al.~\cite{roy2017masterprint} demonstrated the security issues of utilizing a single local patch / partial fingerprint for authentication systems in consumer devices. However, in the proposed approach, the final decision for each image, \textit{i.e.} Spoof or Live, is based on the average Spoofness Score utilizing over 45 minutiae-based local patches.}

\begin{figure*}[t]
\centering
\includegraphics[width=\linewidth]{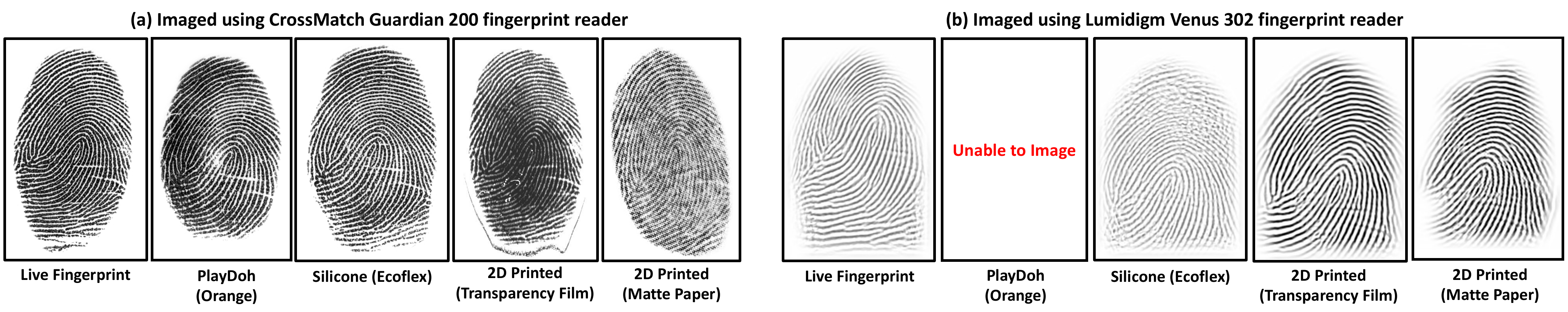}
\caption{Example images from MSU Fingerprint Presentation Attack Dataset (MSU-FPAD) acquired using (a) CrossMatch Guardian 200, and (b) Lumidigm Venus 302 fingerprint readers. Note that Lumidigm reader does not image PlayDoh (orange) spoofs.}
\label{fig:msufpad}
\end{figure*}

\begin{figure*}[t]
\centering
\includegraphics[width=\linewidth]{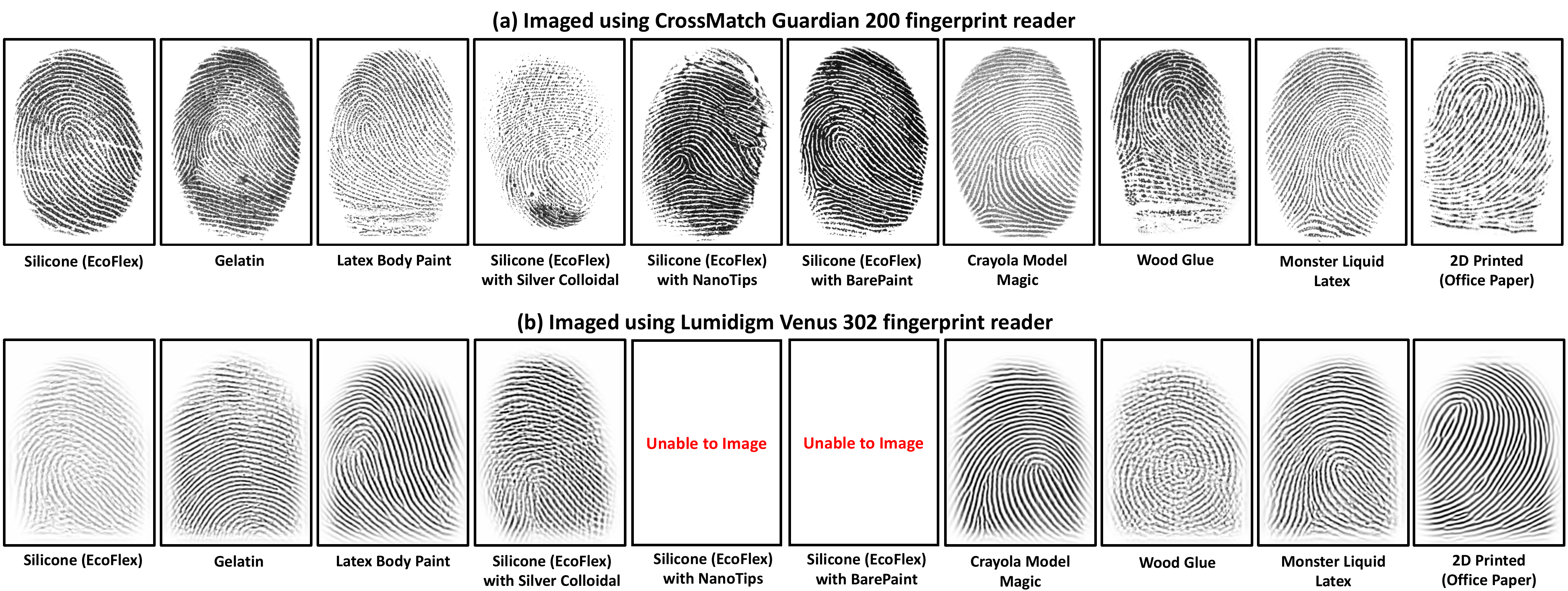}
\caption{Example images from Precise Biometrics Spoof-Kit Dataset (PBSKD) acquired using (a) CrossMatch Guardian 200, and (b) Lumidigm Venus 302 fingerprint readers. Note that Lumidigm reader does not image Silicone (EcoFlex) spoofs with NanoTips and BarePaint coatings.}
\label{fig:pbsk}
\end{figure*}

\section{Proposed Spoof Detection Approach}
The proposed approach includes two stages, an offline training stage and an online testing stage. The offline training stage involves (i)~detecting minutiae in the sensed fingerprint image (live or spoof), (ii)~extracting local patches centered and aligned using minutiae location and orientation, respectively, and (iii)~training MobileNet models on the aligned local patches. During the testing stage, the spoof detection decision is made based on the average of spoofness scores for individual patches output from the MobileNet model. An overview of the proposed approach is presented in Fig.~\ref{fig:overview}.

\subsection{Datasets}
The following datasets have been utilized in this study:
\subsubsection{LivDet Datasets}
In order to evaluate performance of the proposed approach, we utilized LivDet 2011~\cite{yambay2012livdet}, LivDet 2013~\cite{ghiani2013livdet}, and LivDet 2015~\cite{mura2015livdet} datasets. Each of these datasets contains over $16,000$ fingerprint images, acquired from four different fingerprint readers, with equal numbers of live and spoof fingerprints that are equally split between training and testing sets. However, the CrossMatch and Swipe readers from LivDet 2013 dataset were not utilized for evaluation purposes because the (a) LivDet competition organizers found anomalies in the fingerprint data from CrossMatch reader and discouraged its use for comparative evaluations~\cite{ghiani2017review}, and (b) the resolution of fingerprint images output from Swipe reader is very low, \textit{i.e.} 96 dpi. Unlike other LivDet datasets, spoof fingerprint images from Biometrika and Italdata readers in LivDet 2013 dataset~\cite{ghiani2013livdet} are fabricated using the \textit{non-cooperative method} \textit{i.e.} without user cooperation. It should be noted that in LivDet 2015, the testing set included spoofs fabricated using new materials, that were not known in the training set. These new materials included liquid ecoflex and RTV for Biometrika, Digital Persona, and Green Bit readers, and OOMOO and gelatin for Crossmatch reader. Table~\ref{tab:datasets} presents a summary of the LivDet datasets used in this study.%, Table~\ref{tab:performanceLivDet} presents a performance comparison between software-based spoof detection solutions and  and  utilizing these datasets.

\subsubsection{MSU Fingerprint Presentation Attack Dataset}
In addition to utilizing LivDet Datasets, we collected a large dataset, called the MSU Fingerprint Presentation Attack Dataset (MSU-FPAD), using two different fingerprint readers, namely, CrossMatch Guardian 200 and Lumidigm Venus 302. There are a total of $9,000$ live images and $10,500$ spoof images captured using these two readers and 4 different spoof fabrication materials, namely, ecoflex, playhoh, 2D printed on matte paper, and 2D printed on transparency film. The selection of the fingerprint readers and the spoof materials is based on the requirements of IARPA ODIN program\footnote{https://www.iarpa.gov/index.php/research-programs/odin/odin-baa} evaluation. This dataset will be further augmented with additional subjects, spoof fabrication materials, and fingerprint readers. Fig.~\ref{fig:msufpad} presents some example fingerprint images, and Table~\ref{tab:datasetsMSU} presents a summary of the MSU Fingerprint Presentation Attack Dataset.

\subsubsection{Precise Biometrics Spoof-Kit Dataset}
We also collected another dataset$^6$ containing high quality $900$ spoof fingerprint images fabricated using $10$ different types of spoof materials, namely, (i) Ecoflex, (ii) Gelatin, (iii) Latex body paint, (iv) Ecoflex with silver colloidal ink coating, (v) Ecoflex with BarePaint coating, (vi) Ecoflex with Nanotips coating, (vii) Crayola Model Magic, (viii) Wood glue, (ix) Monster Liquid Latex, and (x) 2D printed fingerprint on office paper. The spoof specimens used for this dataset are taken from Precise Biometrics\footnote{\url{https://precisebiometrics.com/}} Spoof-Kit containing $10$ specimens per spoof type, for a total of $100$ spoof specimens. Each spoof specimen is imaged $5$ times using two fingerprint readers, namely, CrossMatch Guardian 200 and Lumidigm Venus 302. Note that Lumidigm reader does not image Silicone (EcoFlex) spoofs with NanoTips and BarePaint coatings. An additional $900$ randomly sampled live fingerprints from MSU-FPAD are selected for a total of $1,800$ fingerprint images in Precise Biometrics Spoof-Kit Dataset. Fig.~\ref{fig:pbsk} presents some example fingerprint images, and Table~\ref{tab:datasetsMSU} presents a summary of the Precise Biometrics Spoof-Kit Dataset.

\subsection{Minutiae Detection}
The fingerprint minutiae are extracted using the algorithm from~\cite{cao2011fingerprint}. The three LivDet datasets (LivDet 2011, 2013, and 2015) comprise of fingerprint images captured at varying resolutions, ranging from $500$ dpi to $1000$ dpi. Since the minutiae detector in~\cite{cao2011fingerprint} was designed for 500 dpi images, all fingerprint images are resized to ensure a standard resolution of 500 dpi. A standard resolution for all the fingerprint images is also crucial to ensure similar amount of friction ridge area in each local patch, irrespective of the fingerprint reader used. An average of 46 minutiae (s.d. = 6.2) and 50 minutiae (s.d. = 6.9) are detected per live image and spoof image, respectively, for the LivDet datasets.

\begin{figure*}[t]
\centering
\includegraphics[width=\linewidth]{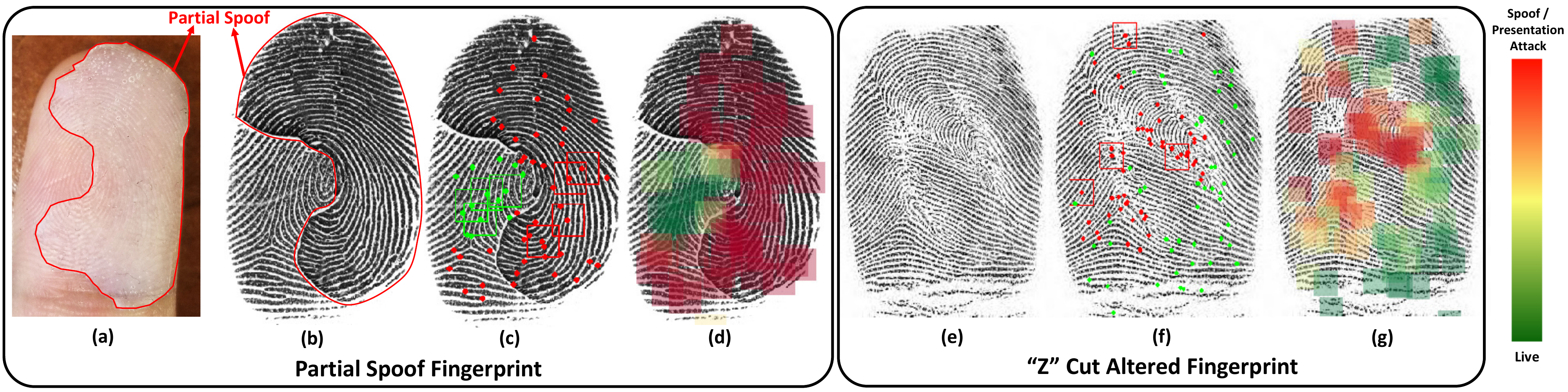}
\caption{The proposed approach provides a fine-grained representation for spoof detection by using minutiae-based local patches. A fingerprint spoof fabricated using silicone which conceals only a partial region of the live finger is shown in (a) and the imaged fingerprint in (b) (enclosed in red). The proposed approach extracts and evaluates the minutiae-based local patches, and highlights the local regions as live (in green) or spoof (in red) as shown in (c) and (d). It can also highlight the regions of fingerprint alterations as shown for a ``Z" cut altered fingerprint in (e), (f) and (g). The proposed approach detected (b) and (e) as spoofs with the spoofness scores of $0.78$ and $0.65$, respectively.}
\label{fig:partial_altered}
\end{figure*}

\begin{figure}[t]
\centering
\includegraphics[trim=0.2cm 0cm 0.2cm 0cm, width=\linewidth]{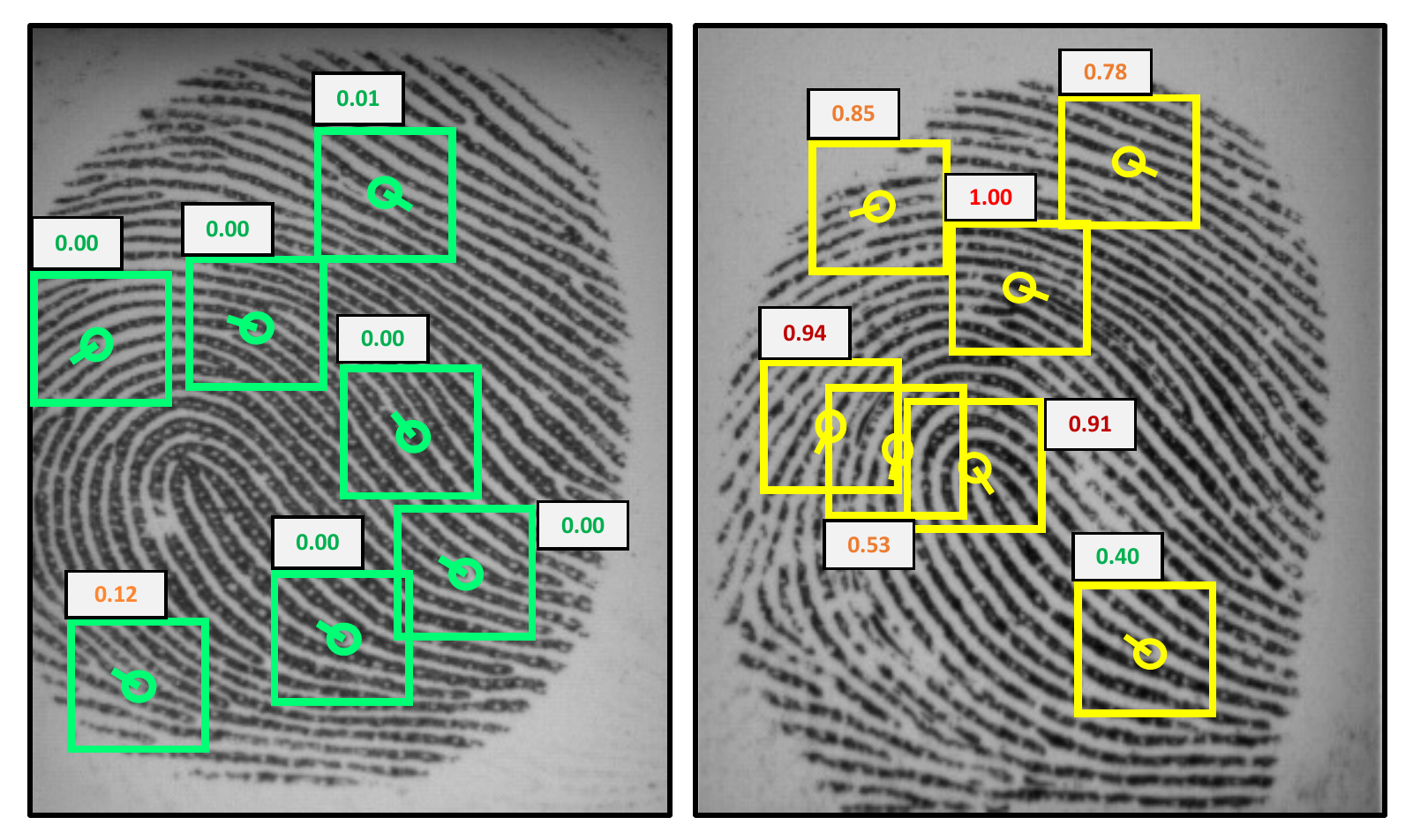}
\caption{Local patches extracted around the fingerprint minutiae for (a) real fingerprint, and (b) spoof fingerprint (gelatin). The spoofness score for each patch is in the range $[0-1]$; higher the score, more likely the patch is extracted from a spoof fingerprint. For a given input test image, the spoofness scores corresponding to the local patches are averaged to give a global spoofness score. The final decision is made based on a classification threshold learned from the training dataset; an image with a global spoofness score below the threshold is classified as live, otherwise as spoof.}
\label{fig:SpoofnessScore}
\vspace{-3.1mm}
\end{figure}

\begin{figure}[t]
\centering
\includegraphics[width=\linewidth]{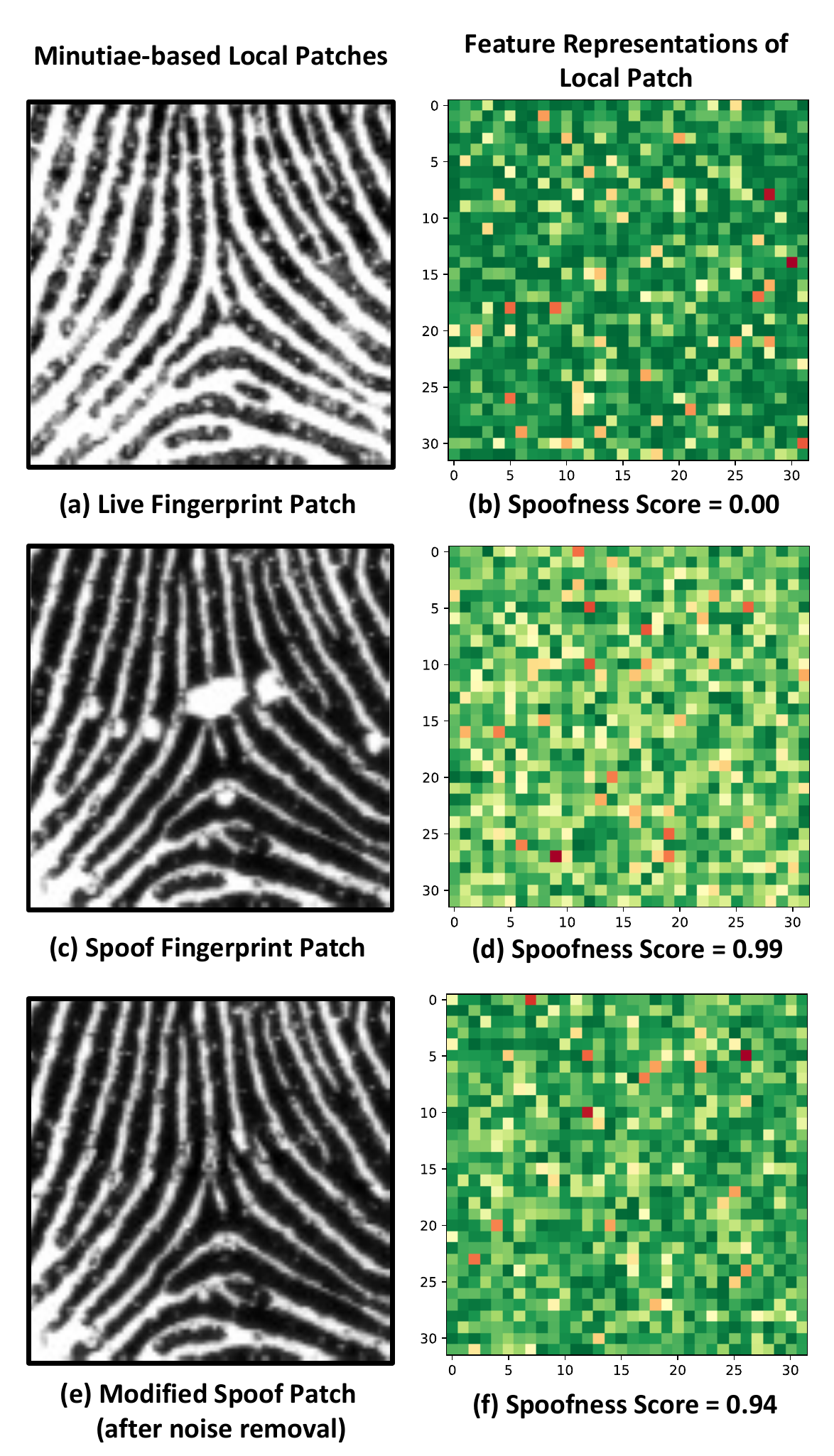}
\caption{Illustrating the embeddings of minutiae-based local patches ($96 \times 96$), for (a) live patch, (c) spoof patch, and (e) modified spoof patch, in 1024-dimensional feature space from MobileNet-v1 bottleneck layer, transformed to $32 \times 32$ heat maps, (b), (d), and (e), respectively, for visualization. A high spoofness score for the modified spoof patch despite removal of artifacts indicates the robustness of the proposed approach.}
\label{fig:outputFeatures}
\end{figure}

\subsection{Local Patch Extraction}
For a given fingerprint image $I$ with $k$ detected minutiae points $M = \{m_1, m_2, \dots, m_k\}$, where $m_i = \{x_i,y_i,\theta_i\}$, \textit{i.e.} the minutiae $m_i$ is defined in terms of spatial coordinates ($x_i$, $y_i$) and orientation ($\theta_i$), a corresponding set of $k$ local patches $L = \{l_1, l_2, \dots, l_k\}$, each of size [$q \times q$] where ($q = \sqrt{2}p$), are extracted. Each local patch $l_i$, centered at the corresponding minutia location ($x_i, y_i$), is aligned\footnote{MATLAB's \textit{imrotate} function with bilinear interpolation is used to rotate the local patch for alignment.} based on the minutiae orientation ($\theta_i$). After alignment, the central region of size [$p \times p$] ($p=96$) is cropped from the rotated patch and used for training the CNN model (see~Fig.~\ref{fig:overview}). The size of larger patch is fixed to [$\sqrt{2}p \times \sqrt{2}p$] to prevent any loss of information during patch alignment. %To handle the minutiae close to the image boundary, the fingerprint image is padded on all sides with a padsize = $\sqrt{2}p/2$ before minutiae detection.  %\textit{i.e.} some region of the local patch lies outside the image region then the patch region is shifted inwards such that it is completely embedded within the fingerprint region, ensuring the size of each patch to be [$p \times p$].
Fig.~\ref{fig:SpoofnessScore} presents examples of real and spoof fingerprint images and the corresponding local patches centered around minutiae points before alignment. 

For evaluating the impact of local patch size on the spoof detection performance, we also explore use of multi-resolution patches of size $p \in \{64, 96, 128\}$ for training independent CNN models and their fusion. All the local patches are resized\footnote{TensorFlow's resize utility with bilinear interpolation was used; available at {\scriptsize\burl{https://www.tensorflow.org/api_docs/python/tf/image/resize_images}}} to $224 \times 224$ as required by the Mobilenet-v1 model.\\

%A convolutional layer applies a set of filters on local regions of the input image to learn a feature map $z$ corresponding to each filter. A non-linear function, $f(z)$, such as ReLU ($max(0,z)$), is applied to each feature map, and the corresponding outputs are then passed to the pooling layer. A pooling layer merges the feature information in local regions by applying a pool operation, such as max-pooling ($p = maxpool(f(z))$), to generate a pooled feature map (usually smaller in size). Max-pooling selects the maximum response in the local regions of the feature map, such that the pooled feature map is less sensitive to slight variations in position, shape, illumination, etc.

\subsection{MobileNet CNN} %Convolutional Neural Network}
%%\vspace{-1mm}
%A Convolutional Neural Network (CNN) is a cascade of multiple layers consisting of linear and non-linear processing units. These layers, when stacked together, can learn a complex multi-level representation of the input data corresponding to different levels of abstraction. 
Since the success of AlexNet~\cite{krizhevsky2012imagenet} for object detection in ILSVRC-2012~\cite{russakovsky2015imagenet}, different CNN architectures have been proposed in literature, such as VGG, GoogleNet (Inception v1-v4), ResNets, MobileNet, etc. Nogueira et al.~\cite{nogueira2016fingerprint}, winner of LivDet 2015, utilized a pre-trained VGG architecture~\cite{simonyan2014very} to achieve the best performance in LivDet 2015~\cite{mura2015livdet}. In this study, we utilize the MobileNet-v1 architecture~\cite{howard2017mobilenets} because it offers the following advantages over VGG and Inception-v3 architectures: (i)~MobileNet-v1 is designed using depth-wise separable convolutions, originally introduced in~\cite{chollet2016xception}, providing drastic decrease in model size and training/evaluation times while providing better spoof detection performance, (ii)~it is a low-latency network requiring only 100ms to classify an input fingerprint image as live or spoof compared to 800ms required by Inception-v3 network~\cite{chugh2017fingerprint}, (iii)~the number of model parameters to be trained in MobileNet-v1 (4.24M), is significantly smaller than the number of model parameters in Inception-v3 (23.2M) and VGG (138M), requiring significantly lower efforts in terms of regularization and data augmentation, to prevent overfitting~\cite{howard2017mobilenets}.

We utilized the TF-Slim library\footnote{\noindent{\scriptsize\url{https://github.com/tensorflow/models/tree/master/research/slim}}} implementation of the MobileNet-v1 architecture. The last layer of the architecture, a $1000$-unit softmax layer (originally designed to predict the $1,000$ classes of ImageNet dataset), was replaced with a $2$-unit softmax layer for the two-class problem, \textit{i.e.} live vs. spoof. The optimizer used to train the network was RMSProp with asynchronous gradient descent and a batch size of $100$. Data augmentation techniques, such as brightness adjustment, random cropping, vertical flipping are employed to ensure the trained model is robust to the possible variations in fingerprint images. For the multi-resolution local patches, a separate network is trained for each patch size with the same parameters as mentioned above.

\subsection{Fine-grained Fingerprint Image Representation}
Partial spoofs and fingerprint alterations are not a rare means to avoid re-identification\footnote{\burl{http://abcnews.go.com/Technology/GadgetGuide/surgically-altered-fingerprints- woman-evade-immigration/story?id=9302505}}, by masking the true identity from a fingerprint biometric system~\cite{yoon2012altered,roy2017masterprint}. Spoof detectors trained on the whole fingerprint images are inadvertently ineffective against localizing partial spoof fingerprints, that conceal only a limited region of the live finger.  Moreover, in many smartphones and other embedded systems that only sense a partial region (friction ridge area) of the fingerprint due to small sensor area (typically $150 \times 150$), it is very crucial to have a detailed representation of the sensed fingerprint region. One of the key advantages of employing a patch-based approach is the fine-grained representation of input fingerprint image for spoof detection. Fig.~\ref{fig:partial_altered}~(a) presents an example of a fingerprint spoof fabricated using silicone, concealing only a partial region of the live finger and Fig.~\ref{fig:partial_altered}~(b) presents the imaged partial spoof fingerprint using a CrossMatch Guardian 200 fingerprint reader. The proposed approach, utilizing minutiae-based local patches, highlights the local regions as live or spoof (shown in Figs.~\ref{fig:partial_altered}~(c) and (d) in green and red, respectively), providing a fine-grained representation of the fingerprint image. Fingerprint alterations, such as cuts, mutilations, stitches, etc., performed using surgical and chemical procedures (see Fig.~\ref{fig:partial_altered}~(e)), create spurious minutiae as shown in Figs.~\ref{fig:partial_altered} (f) and (g). The proposed approach is able to highlight the regions of fingerprint alterations despite not being trained specifically on altered fingerprint database, indicating the generalizability of the proposed approach. The proposed approach detected both fingerprint images in Figs.~\ref{fig:partial_altered} (b) and (e) as spoofs with the spoofness scores of $0.78$ and $0.65$, respectively.

\subsection{Spoofness Score}
The output from the softmax layer of the trained MobileNet-v1 model is in the range [$0-1$], called as the \textit{Spoofness Score}. The larger the spoofness score, the higher the support that the input local patch belongs to the spoof class (see Fig.~\ref{fig:SpoofnessScore}). For an input test image $I$, the spoofness scores $s_{i \in \{1,2,...,k\}}^I$ corresponding to the $k$ minutiae-based local patches of size $p \times p$, extracted from the input image, are averaged to give a global spoofness score $S^I$. In case of multi-resolution local patches, the global spoofness scores ($S^I_{p_i}$) based on each local patch size, $p_i \in \{64,96,128\}$, are averaged to produce a final spoofness score. The threshold that minimizes the average classification error on training dataset is learned and utilized as the classification threshold. An image with a spoofness score below the threshold is classified as live, otherwise as spoof. The learned threshold performed slightly better in spoof detection than selecting a pre-defined threshold of $0.5$.
%\subsection{Random Patches vs. Minutiae-based Patches}

\subsection{On Robustness of Patch-based Representation}
As mentioned in Section-1, the proposed patch-based approach addresses several drawbacks of using whole fingerprint images for training deep CNN architectures. While the proposed approach is based on the premise that it is capable of capturing discriminatory information from local patches, such as missing ridges, air bubbles, etc., from spoof fingerprints, we also examine the robustness of patch-based representation by evaluating it in absence of such artifacts. Figs.~\ref{fig:outputFeatures} (a) and (c) present minutiae-based local patches from a live fingerprint and the corresponding spoof fingerprint (fabricated using EcoFlex), respectively, for the same minutia point. Figs.~\ref{fig:outputFeatures} (b) and (d) present the feature representations of the local patches shown in Figs.~\ref{fig:outputFeatures} (a) and (c), respectively, obtained from the bottleneck layer of the MobileNet-v1 architecture. The 1024-dimensional feature representation is transformed to $32 \times 32$ heatmap for visualization. The spoofness scores for the two patches are $0.00$ (Fig.~\ref{fig:outputFeatures} (b)) and $0.99$ (Fig.~\ref{fig:outputFeatures} (d)), respectively. The spoof patch (Fig.~\ref{fig:outputFeatures} (c)) is modified, by the authors, using an open-source photo-editing utility, called \textit{GIMP}\footnote{\url{https://www.gimp.org/}}, to remove the visible artifacts and produce the modified spoof fingerprint patch as shown in Fig.~\ref{fig:outputFeatures} (e). The feature representation for the modified patch is shown in Fig.~\ref{fig:outputFeatures} (f). A high spoofness score for the modified spoof patch ($0.94$)  despite removal of artifacts indicates the robustness of the proposed approach.

\section{Experimental Results}

\subsection{Performance Evaluation Metrics}
The performance of the proposed approach is evaluated following the metrics used in LivDet~\cite{ghiani2017review}.

\begin{itemize}
\item $F${\footnotesize$errlive$}: Percentage of misclassified live fingerprints.
\item $F${\footnotesize$errfake$}\footnote{When all the spoof fabrication materials are known during the training, this metric is referred to as $F${\footnotesize$errfake\_known$}, and in case all the spoof fabrication materials to be encountered during testing are not known during training, this metric is referred to as $F${\footnotesize$errfake\_unknown$}.}: Percentage of misclassified spoof fingerprints. 
\end{itemize}
The average classification error ($ACE$) is defined as:
\begin{equation}
ACE = \frac{F_{errlive}+F_{errfake}}{2}
\end{equation}

Additionally, we also report the $F${\footnotesize$errfake$} @ $F${\footnotesize$errlive$}$~= 1\%$ for each of the experiments as reported in~\cite{ghiani2017review}. This value represents the percentage of spoofs able to breach the biometric system security when the reject rate of legitimate users $\le 1\%$.

%% LivDet 2015
\begin{table*}[t!]
\centering
\caption{Performance comparison between the proposed approach (bottom) and state-of-the-art (top) reported on LivDet 2015 dataset~\cite{mura2015livdet}. Separate networks are trained on the training images captured by each of the four fingerprint readers. \textit{Ferrfake known} and \textit{Ferrfake unknown} correspond to Known Spoof Materials and Cross-Material scenarios, respectively.}
\label{tab:intrasensor1}
\resizebox{\textwidth}{!}{
\begin{tabu}{ | p{0.25 cm} | p{1.8cm} | >{\centering\arraybackslash}p{1.8cm} | >{\centering\arraybackslash}p{1.8cm} | >{\centering\arraybackslash}p{1.7cm} | >{\centering\arraybackslash}p{2cm} | >{\centering\arraybackslash}p{1.3cm} | >{\centering\arraybackslash}p{2.4cm} |}
\hline
%\textbf{Dataset} & \multicolumn{2}{ c |}{\textbf{LivDet 2015 Winner}~\cite{mura2015livdet}} & \multicolumn{3}{ c |}{\textbf{Proposed Approach}} \\ \hline
 \multirow{4}{*}{\rot{\textbf{{\footnotesize State-of-the-Art~\cite{mura2015livdet}}}}} & \textbf{LivDet 2015} & \textbf{Ferrlive (\%)} & \textbf{Ferrfake$^\dagger$ (\%)}  & \textbf{Ferrfake known (\%)} & \textbf{Ferrfake unknown* (\%)} & \textbf{ACE (\%)} & \textbf{Ferrfake (\%) @ Ferrlive= 1\%~\cite{ghiani2017review}} \\ \cline{2-8}
 & GreenBit           & 3.50 & 5.33 & 4.30 & 7.40 & 4.60 & 17.90 \\ \cline{2-8}
& Biometrika         & 8.50 & 3.73 & 2.70 & 5.80 & 5.64 & 15.20  \\ \cline{2-8}
& Digital Persona  & 8.10 & 5.07 & 4.60 & 6.00 & 6.28 & 19.10 \\ \cline{2-8}
& Crossmatch       & 0.93 & 2.90 & 2.12 & 4.02 & 1.90 & 2.66 \\ \cline{2-8}
& \textbf{Average} & 4.78 & 4.27 & 3.48 & 5.72 & 4.49 & 13.24\\ \hline
& & & & & & & \\
\hline
\multirow{4}{*}{\rot{\textbf{{\footnotesize Proposed Approach}}}} & \textbf{LivDet 2015} & \textbf{Ferrlive (\%)} & \textbf{Ferrfake$^\dagger$ (\%)}  & \textbf{Ferrfake known (\%)} & \textbf{Ferrfake unknown* (\%)} & \textbf{ACE (\%)} & \textbf{Ferrfake (\%) @ Ferrlive = 1\%} \\ \cline{2-8}
 & GreenBit           & 0.50 & 0.80 & 0.30 & 1.80 & \textbf{0.68} & 0.53 \\ \cline{2-8}
& Biometrika         & 0.90 & 1.27 & 0.60 & 2.60 & \textbf{1.12} & 1.20 \\ \cline{2-8}
& Digital Persona  & 1.97 & 1.17 & 0.85 & 1.80 & \textbf{1.48} & 1.96 \\ \cline{2-8}
& Crossmatch       & 0.80 & 0.48 & 0.82 & 0.00 & \textbf{0.64} & 0.28 \\ \cline{2-8}
& \textbf{Average} & \textbf{1.02} & \textbf{0.93} & \textbf{0.64} & \textbf{1.48} & \textbf{0.97} & \textbf{0.96}\\ \hline
\end{tabu}
}
%\vspace{-2.8mm}
\begin{flushleft}
{\footnotesize $\dagger$ Ferrfake includes spoofs fabricated using both known and previously unseen materials. It is an average of Ferrfake-known and Ferrfake-unknown, weighted by the number of samples in each category.\\}
{\footnotesize *The unknown spoof materials in LivDet 2015 test dataset include Liquid Ecoflex and RTV for Green Bit, Biometrika, and Digital Persona sensors, and OOMOO and Gelatin for Crossmatch sensor.\\}
\end{flushleft}
\vspace{-2.8mm}
\end{table*}

%% LivDet 2011 2013
\begin{table}[t]
\centering
\caption{Performance comparison between the proposed approach and state-of-the-art results reported on LivDet~2011 and LivDet~2013 datasets for intra-sensor experiments in terms of Average Classification Error (ACE) and Ferrfake~@~Ferrlive~=~1\%.}
\label{tab:intrasensor2}
\resizebox{\linewidth}{!}{
\begin{tabular}{ | p{1.78 cm} || >{\centering\arraybackslash}p{2cm} ||  >{\centering\arraybackslash}p{1.3cm} | >{\centering\arraybackslash}p{1.8 cm} | }
\hline
\textbf{Dataset} & \textbf{State-of-the-Art} & \multicolumn{2}{ c |}{\textbf{Proposed Approach}} \\ \hline

\textbf{LivDet 2011} & \textbf{ACE (\%)} & \textbf{ACE (\%)} & \textbf{Ferrfake @ Ferrlive = 1\%}  \\ \hline
Biometrika &  4.90~\cite{gragnaniello2015local} & \textbf{1.24}  & 1.41 \\ \hline
Digital Persona &  1.85~\cite{pala2017deep} & \textbf{1.61}  & 3.25 \\ \hline
ItalData &  5.10~\cite{pala2017deep} & \textbf{2.45} & 7.21 \\ \hline
Sagem &  \textbf{1.23}~\cite{pala2017deep} & 1.39  & 4.33 \\ \hline
\textbf{Average} & 3.27 & \textbf{1.67} & 4.05 \\ 
\hline \hline

 \textbf{LivDet 2013} & & & \\ \hline
Biometrika & 0.65~\cite{pala2017deep} & \textbf{0.20} & 0.00  \\ \hline
ItalData & 0.40~\cite{nogueira2016fingerprint} & \textbf{0.30} & 0.10 \\ \hline
\textbf{Average} & 0.53 & \textbf{0.25} & 0.05 \\ \hline

\end{tabular}
}
\end{table}

%% MSU FPAD and PBSK
\begin{table}[t]
\centering
\caption{Average Classification Error (ACE), Ferrfake~@~Ferrlive~=~0.1\% and Ferrlive~=~1\% on the MSU Fingerprint Presentation Attack Dataset (MSU-FPAD) and Precise Biometrics Spoof-Kit Dataset (PBSKD) for intra-sensor experiments.}
\label{tab:intrasensor_msupbsk}
\resizebox{\linewidth}{!}{
\begin{tabular}{ | p{3 cm} || >{\centering\arraybackslash}p{1.3cm} |  >{\centering\arraybackslash}p{2cm} | >{\centering\arraybackslash}p{1.8 cm} | }
\hline
\textbf{Dataset} & \multicolumn{3}{ c |}{\textbf{Proposed Approach}} \\ \hline
\textbf{MSU-FPAD} & \textbf{ACE (\%)} & \textbf{Ferrfake @ Ferrlive = 0.1\%} & \textbf{Ferrfake @ Ferrlive = 1\%}  \\ \hline
CrossMatch Guardian 200 & 0.08 & 0.11 & 0.00 \\ \hline
Lumidigm Venus 302 & 3.94 & 10.03  & 1.30 \\ \hline
\textbf{Average} & \textbf{2.01} & \textbf{5.07} & \textbf{0.65} \\
\hline \hline

 \textbf{PBSKD} & & & \\ \hline
CrossMatch Guardian 200 & 2.02 & 5.32 & 0.65 \\ \hline
Lumidigm Venus 302 & 1.93 & 3.84 & 0.33 \\ \hline
\textbf{Average} & \textbf{1.98} & \textbf{4.66} & \textbf{0.51} \\ \hline
\end{tabular}
}
\end{table}

\subsection{Results}
The proposed approach is evaluated under the following four scenarios of fingerprint spoof detection, which reflect an algorithm's robustness against new spoof materials, use of different sensors and/or different environments.

\subsubsection{Intra-Sensor, Known Spoof Materials}

\begin{figure*}[t]
\centering
%%\vspace{-1mm}
\includegraphics[trim=0cm 0cm 0cm 0cm, width=0.92\linewidth]{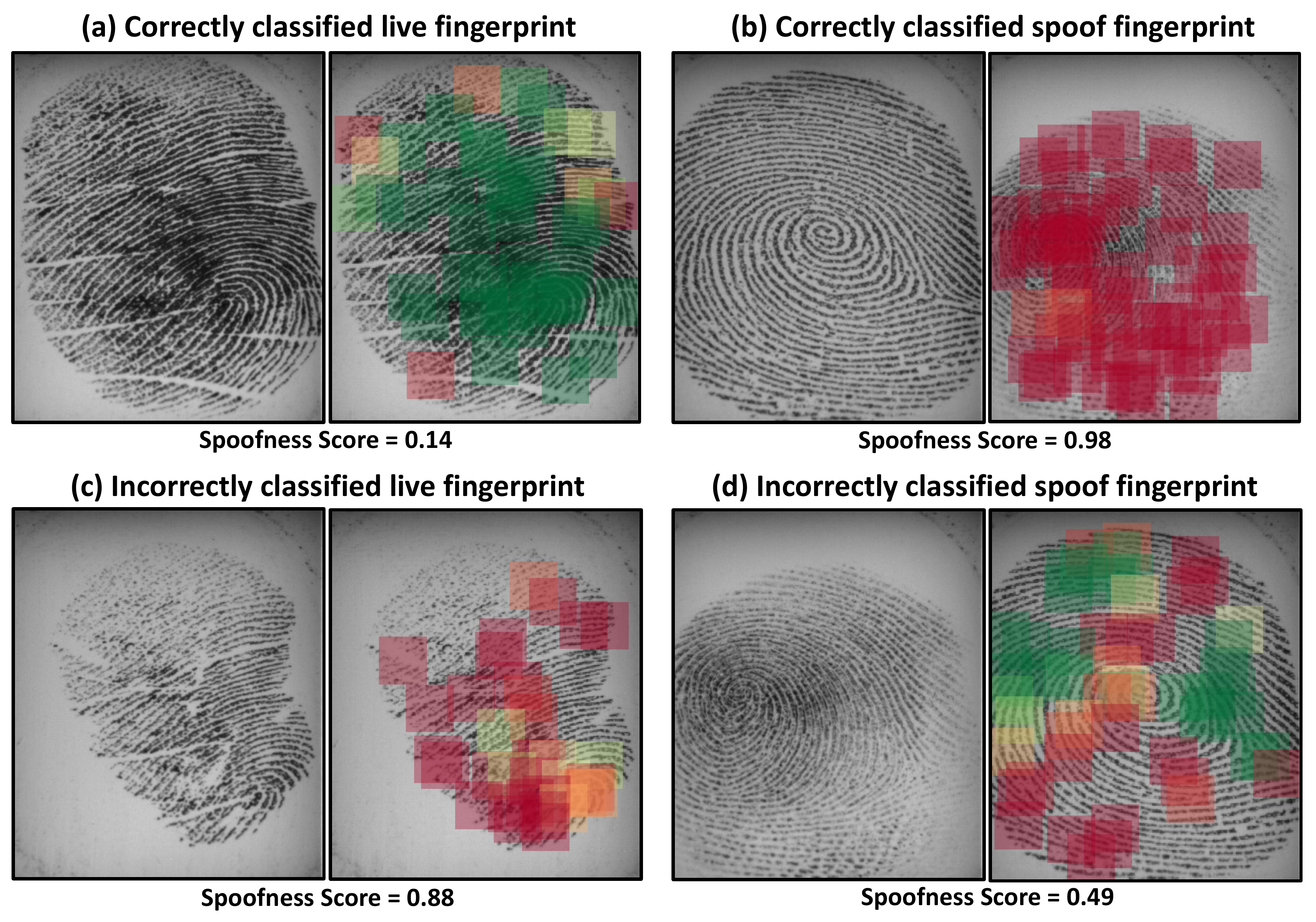}
\caption{Example live and spoof fingerprints for Biometrika sensor from LivDet 2015 dataset, correctly and incorrectly classified by our proposed approach.}
\label{fig:examples}
\vspace{-4mm}
\end{figure*}

In this setting, all the training and testing images are captured using the same sensor, and all spoof fabrication materials utilized in the test set are known a priori. Our experimental results show that training the MobileNet-v1 model from scratch, using minutiae-based local patches, performs better than fine-tuning a pre-trained network, as reported in~\cite{nogueira2016fingerprint}. The large amount of available data, in the form of local fingerprint patches, is sufficient to train the deep architecture of MobileNet-v1 model without over-fitting. Also, a score level fusion of the proposed MobileNet-v1 model trained on local patches, with an independent MobileNet-v1 model trained on whole fingerprint images does not offer any advantage in terms of performance improvement. This observation was also made in our preliminary study~\cite{chugh2017fingerprint} which employed Inception-v3 model~\cite{szegedy2016rethinking}. % as the accuracy of the full image based model was found to be much lower. %, unlike inwhere the VGG model pre-trained on ImageNet  and fine-tuned using whole fingerprint images, our  performed better when trained from scratch using the s. %This observation can be attributed to two plausible reasons: (i)~Inception-v3 has a relatively deeper architecture compared to VGG and AlexNet; the complex multi-level representations learnt by this deep architecture on large ImageNet  might be too strong to be fine-tuned, and (ii)~the large amount of available fingerprint data, in the form of local patches, is sufficient to train the deep architecture without over-fitting. We also tried  For instance, fusion on LivDet 2015 dataset increased the average classification error from 1.39\% (for minutiae-based approach) to 3\%. 
It was reported in~\cite{ghiani2017review} that most of the algorithms submitted to LivDet 2015 did not perform well on Digital Persona sensor due to the small image size. Our approach based on local patches does not suffer from this limitation. Tables~\ref{tab:intrasensor1} and~\ref{tab:intrasensor2} present the performance comparison between the proposed approach and the state-of-the-art results for the LivDet datasets utilized in this study. Table~\ref{tab:intrasensor_msupbsk} presents the performance of the proposed approach on MSU Fingerprint Presentation Attack Dataset (MSU-FPAD) and Precise Biometrics Spoof-Kit Dataset (PBSKD). Independent MobileNet-v1 networks are trained for each evaluation. Note that in LivDet 2015 (Table~\ref{tab:intrasensor1}), this scenario is represented by the \textit{Ferrfake known}. For  LivDet 2011 and 2013, MSU-FPAD, and PBSKD datasets (Table~\ref{tab:intrasensor2}), all spoof materials in the test set were known during training. Fig.~\ref{fig:examples} presents example fingerprint images for Biometrika sensor from LivDet 2015 dataset that were correctly and incorrectly classified by the proposed approach.
%The spoof fingerprints for LivDet 2013 Biometrika and ItalData sensors were fabricated using the non-cooperative method (\textit{i.e.} without user cooperation), which explains the near perfect accuracy on these datasets.

%%Biometrika Examples

\begin{figure}[t]
\centering
\includegraphics[width=\linewidth, trim=0cm 6cm 0.7cm 5cm]{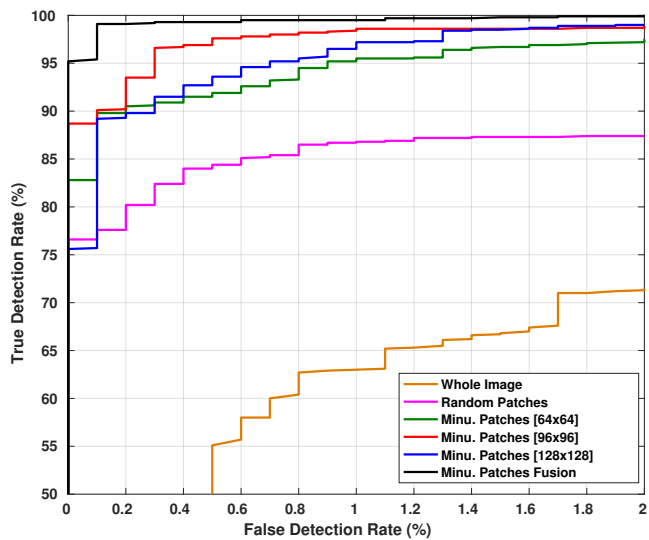}
\caption{ROC curves for live v. spoof classification of fingerprint images from LivDet 2011 Dataset (Biometrika sensor) utilizing (i) whole image, (ii) randomly selected patches [$96 \times 96$], (iii) minutiae-based patches of size [$p \times p$], $p \in \{64,96,128\}$, (iv) score-level fusion of multi-resolution patches.}
\label{fig:patchSize}
\vspace{-2mm}
\end{figure}

We also evaluate the impact of local patch size on the performance of the proposed approach, by comparing the performance of three CNN models trained on minutiae-centered local patches of size $[p \times p]$ where $p = \{ 64, 96,128\}$, extracted from the fingerprint images captured by Biometrika sensor for LivDet 2011 dataset. Among these three models, the one trained on local patches of size [$96 \times 96$] performed the best. However, a score-level fusion, using average-rule, of the three models reduced the average classification error (ACE) from $1.24\%$ to $0.88\%$, and Ferrfake from $1.41\%$ to $0.58\%$ @ Ferrlive $= 1\%$. Similar performance gains were observed for other sensors, but there is a trade off between the performance gain and the computational requirements for spoof detector. In order to evaluate the significance of utilizing minutiae locations for extracting local patches, we trained independent MobileNet-v1 models on a similar number of local patches, extracted randomly from LivDet 2015 datasets. It was observed that the models trained on minutiae-centered local patches achieved a significantly higher reduction ($78\%$) in average classification error, compared to the reduction ($33\%$) achieved by the models trained on randomly sampled local patches. Fig.~\ref{fig:patchSize} illustrates that (i) features extracted from local patches provide better spoof detection accuracy than the whole image, (ii) patches selected around minutiae perform better than random patches of the same size, (iii) $96 \times 96$ patch performs the best among the three patch sizes considered, and (iv) score-level fusion of multi-resolution local patches boosts the spoof detection performance.

\subsubsection{Intra-Sensor, Cross-Material}

In this setting, the same sensor is used to capture all training and testing images, but the spoof images in the testing set are fabricated using new materials that were not seen during training. For the first set of cross-material experiments, we utilize the LivDet 2015 dataset which contains two new spoof materials in the testing set for each \hbox{sensor}, \textit{i.e.} Liquid Ecoflex and RTV for Green Bit, Biometrika, and Digital Persona sensors, and OOMOO and Gelatin for Crossmatch sensor. The performance of the proposed approach on cross-material experiments for LivDet 2015 dataset is presented in Table~\ref{tab:intrasensor1} (column $F${\footnotesize$errfake\_unknown$}) and is compared with the state-of-the-art performance reported in~\cite{mura2015livdet}. A significant reduction in the error rate is achieved by the proposed method. For better generalizability, a second set of cross-material experiments are performed on LivDet 2011 and LivDet 2013 datasets, following the protocol adopted by the winner of LivDet 2015~\cite{nogueira2016fingerprint}. Table~\ref{tab:crossmaterial} presents the achieved error rates on these experiments, along with the spoof fabrication materials used in training and testing sets.

\begin{table*}[!htbp]
\centering
\caption{Performance comparison between the proposed approach and state-of-the-art results reported on LivDet 2011 and LivDet 2013 datasets for cross-material experiments, in terms of Average Classification Error (ACE) and Ferrfake @ Ferrlive = 1\%.}
\label{tab:crossmaterial}
\resizebox{\textwidth}{!}{
\begin{tabular}{ | p{2 cm} | p{2.9 cm} | p{3.1 cm} | >{\centering\arraybackslash}p{2.2 cm} | >{\centering\arraybackslash}p{1.7 cm} | >{\centering\arraybackslash}p{3.3 cm} | }
\hline
\textbf{Dataset} & \multicolumn{2}{c|}{\textbf{Spoof Materials}} & \textbf{State-of-the-Art} & \multicolumn{2}{ c |}{\textbf{Proposed Approach}} \\ \hline
 & \textbf{Materials - Training} & \textbf{Materials - Testing} & \textbf{ACE (\%)} & \textbf{ACE (\%)} & \textbf{\textbf{Ferrfake @ Ferrlive = 1\%}} \\ \hline
Biometrika 2011 & EcoFlex, Gelatine, Latex & Silgum, WoodGlue & 10.10~\cite{nogueira2016fingerprint} & \textbf{4.60} & 8.15 \\ \hline
Biometrika 2013 & Modasil, WoodGlue & EcoFlex, Gelatine, Latex  & 2.10~\cite{pala2017deep} & \textbf{1.30} & 0.34 \\ \hline
ItalData 2011 & EcoFlex, Gelatine, Latex & Silgum, WoodGlue, Other & 7.00~\cite{pala2017deep} & \textbf{5.20} &  7.80 \\ \hline
ItalData 2013 & Modasil, WoodGlue & EcoFlex, Gelatine, Latex & 1.25~\cite{pala2017deep} & \textbf{0.60} & 0.68 \\ \hline
\multicolumn{3}{|l|}{\textbf{Average}} & 5.11 & \textbf{2.93} & 4.24 \\ \hline
\end{tabular}
}
\vspace{2mm}
\end{table*}

\subsubsection{Cross-Sensor Evaluation}

In this evaluation, the training and the testing images are obtained from two different sensors but from the same dataset. This setting reflects the algorithm's strength in learning the common characteristics used to distinguish live and spoof fingerprints across fingerprint acquisition devices. For instance, using LivDet 2011 dataset, images from Biometrika sensor are used for training, and the images from ItalData sensor are used for testing. We follow the protocol for selection of training and testing sets for cross-sensor and cross-dataset experiments as adopted by Nogueira et al.~\cite{nogueira2016fingerprint}. Table~\ref{tab:crosssensor} compares the average classification error and Ferrfake @ Ferrlive = 1\% for the proposed approach with the results obtained by~\cite{nogueira2016fingerprint} on cross-sensor experiments.

%% CROSS_SENSOR
\begin{table}[htbp]
\centering
\caption{Performance comparison between the proposed approach and state-of-the-art results~\cite{nogueira2016fingerprint} reported on LivDet 2011 and LivDet 2013 datasets for cross-sensor experiments, in terms of Average Classification Error (ACE), and Ferrfake @ Ferrlive = 1\%.}
\label{tab:crosssensor}
\resizebox{\linewidth}{!}{
\begin{tabular}{ | p{2.1cm} | >{\centering\arraybackslash}p{2cm} | >{\centering\arraybackslash}p{1.2cm} | >{\centering\arraybackslash}p{2cm} | }
\hline
\textbf{Training Dataset (Testing Dataset)} & \textbf{State-of-the-Art} & \multicolumn{2}{c|}{\textbf{Proposed Approach}} \\ \hline
& \textbf{ACE (\%)} & \textbf{ACE (\%)} & \textbf{Ferrfake (\%) @ Ferrlive = 1\%} \\ \hline
Biometrika 2011 (ItalData 2011) & 29.35~\cite{pala2017deep} & \textbf{25.35} & 50.81 \\ \hline
ItalData 2011 (Biometrika 2011) & 27.65~\cite{pala2017deep} & \textbf{25.21} & 76.20 \\ \hline
Biometrika 2013 (ItalData 2013) & \textbf{1.5}~\cite{pala2017deep} & 4.30 & 12.73 \\ \hline
ItalData 2013 (Biometrika 2013) & \textbf{2.30}~\cite{nogueira2016fingerprint} & 3.50 & 70.35 \\ \hline
\textbf{Average} & 15.2  & \textbf{14.59} & \textbf{52.52} \\ \hline
\end{tabular}
}
\vspace{2mm}
\end{table}

%% CROSS_DATASET
\begin{table}[t]
\centering
\caption{Performance comparison between the proposed approach and state-of-the-art results~\cite{pala2017deep} reported on LivDet 2011 and LivDet 2013 datasets for cross-dataset experiments, in terms of Average Classification Error (ACE) and Ferrfake @ Ferrlive = 1\%.}
\label{tab:crossdataset}
\resizebox{\linewidth}{!}{
\begin{tabular}{ | p{2.1cm} | >{\centering\arraybackslash}p{2cm} | >{\centering\arraybackslash}p{1.2cm} | >{\centering\arraybackslash}p{2cm} | }
\hline
\textbf{Training Dataset (Testing Dataset)} & \textbf{State-of-the-Art} & \multicolumn{2}{c|}{\textbf{Proposed Approach}} \\ \hline
& \textbf{ACE (\%)} & \textbf{ACE (\%)} & \textbf{Ferrfake (\%) @ Ferrlive = 1\%} \\ \hline
Biometrika 2011 (Biometrika 2013) & 14.00~\cite{pala2017deep} & \textbf{7.60} & 89.60 \\ \hline
Biometrika 2013 (Biometrika 2011) & 34.05~\cite{pala2017deep} & \textbf{31.16} & 78.84 \\ \hline
ItalData 2011 (ItalData 2013) & 8.30~\cite{pala2017deep} & \textbf{6.70} & 16.70 \\ \hline
ItalData 2013 (ItalData 2011) & 44.65~\cite{pala2017deep} & \textbf{26.16} & 75.09 \\ \hline
\textbf{Average} & 25.25 & \textbf{17.91} & 65.06 \\ \hline
\end{tabular}
}
\end{table}

\begin{figure*}[t]
\centering
\includegraphics[ width=\linewidth]{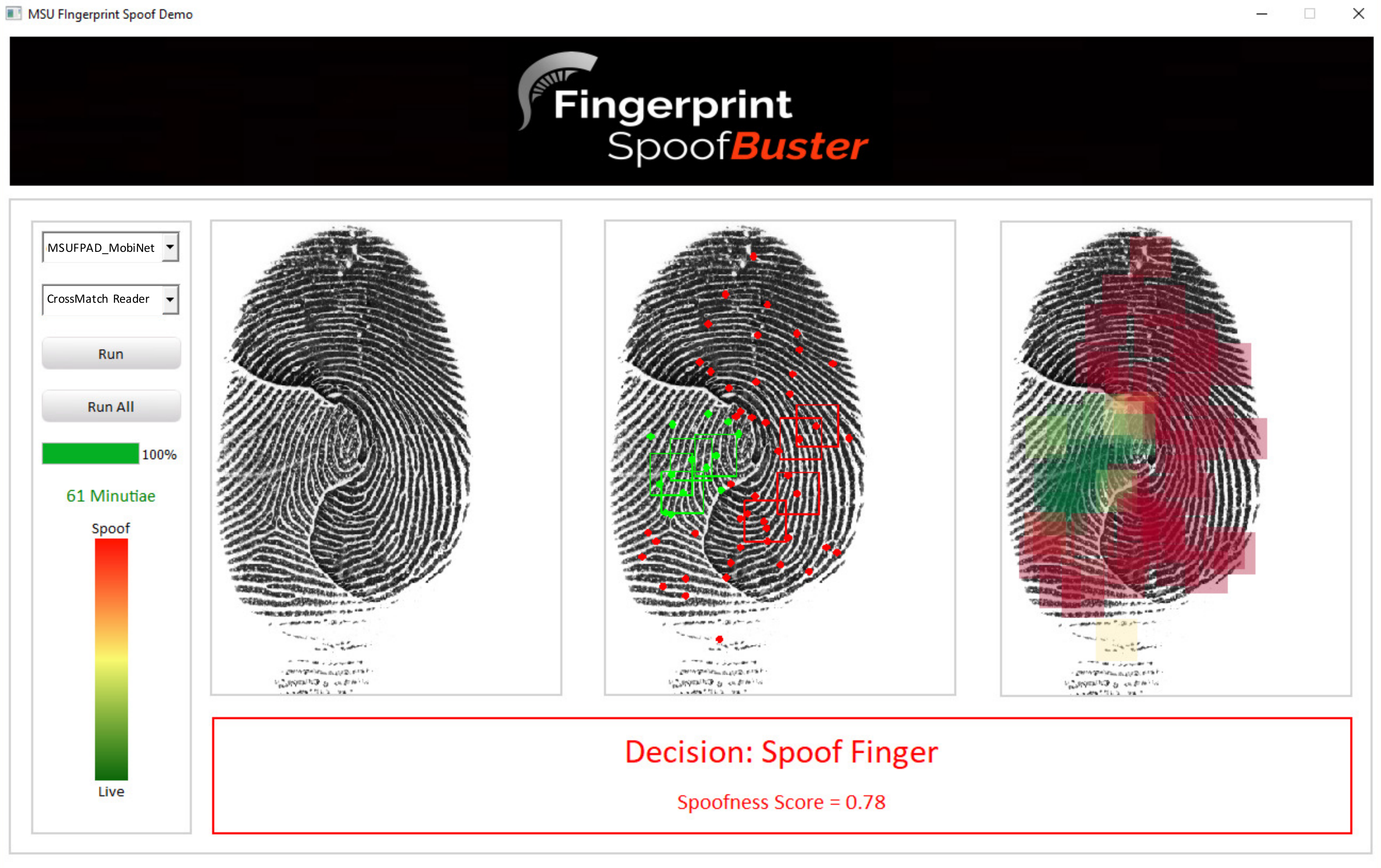}
\caption{Interface of the real-time fingerprint spoof detector, called Fingerprint Spoof Buster. It allows selection of the fingerprint reader and CNN model.}
\label{fig:spoofDemo}
\end{figure*}

\subsubsection{Cross-Dataset Evaluation}

In this scenario, the training and the testing images are obtained using the same sensor, but from two different datasets. For instance, training images are acquired using Biometrika sensor from LivDet 2011 dataset and the testing images are acquired using Biometrika sensor from LivDet 2013. This set of experiments captures the algorithm's invariance to the changes in environment for data collection. Table~\ref{tab:crossdataset} presents the average classification error and Ferrfake @ Ferrlive = 1\%. Results in Table~\ref{tab:crossdataset} show that the proposed local patch based approach achieves a reduction of 29\% in the average classification error from 25.25\% in~\cite{pala2017deep} to 17.91\% in our approach. However, the average Ferrfake @ Ferrlive = 1\% that we report is 52.52\% and 65.06\% for cross-sensor and cross-dataset scenarios respectively, indicating the challenges, especially in applications where a high level of spoof detection accuracy is needed.

\subsection{Processing Times}

The MobileNet-v1 CNN model takes around 6-8 hours to converge using a single Nvidia GTX 1080 Ti GPU utilizing approximately $96,000$ local patches for a training set with $2,000$ fingerprint images ($2,000$ images $\times$ $48$~patches/fingerprint image). The average classification time for a single input image, including minutiae detection, local patch extraction and alignment, inference of Spoofness Scores for local patches, and producing the final decision, on a single Nvidia GTX 1080 Ti GPU is $100$ms.

%\subsection{Improvements due to Multi-Scale Patches}
%\begin{figure}[t]
%\centering
%\includegraphics[trim=4cm 6.9cm 4.7cm 8cm, width=0.76\linewidth]{roc.pdf}
%\caption{Receiver Operating Characteristic (ROC) curve for the models trained on multi-scale local patches of size [$p \times p$], $p = \{ 64, 96, 128\}$, and their fusion.}
%\label{fig:multiscale}
%\end{figure}
%As shown in Table~\ref{tab:intrasensor2}, the proposed approach achieved a poor performance of Fcorrlive = 51.90\% @ Fcorrfake = 99.9\% for Biometrika sensor from LivDet 2011. 

\section{Fingerprint Spoof Buster}
A graphical user interface for real-time fingerprint spoof detection, called Fingerprint Spoof Buster\footnote{This interface will be made available to interested researchers once this paper is accepted for publication.}, is developed. This interface allows the operator to select a specific fingerprint reader and a trained MobileNet-v1 model for evaluation. The operator can perform the evaluation in either \textit{online} or \textit{batch} mode. In the \textit{online} mode, a fingerprint is imaged using the selected reader and displayed on the interface (see Fig.~\ref{fig:spoofDemo}). The extracted fingerprint minutiae and the corresponding local patches are presented and color coded based on their respective spoofness scores (green for live and red for spoof). The global spoofness score and the final decision for the input image is also presented on the interface. In the batch mode, all fingerprint images within a specified directory are evaluated, and global spoofness scores for each file are output in a file. The graphical user interface allows the operator to visually examine the local regions of the fingerprint highlighted as live or spoof, instead of relying on only a single score as output by the traditional approaches. In the future, this interface will be extended to display output from multiple CNN models for an easy visual comparison.

\section{Conclusions}
A robust and accurate method for fingerprint spoof detection is critical to ensure the reliability and security of the fingerprint authentication systems. In this study, we have utilized fingerprint domain knowledge by extracting local patches centered and aligned using minutiae in the input fingerprint image for training MobileNet-v1 CNN models. The local patch based approach provides salient cues to differentiate spoof fingerprints from live fingerprints. The proposed approach is able to achieve a significant reduction in the error rates for intra-sensor (63\%), cross-material (43\%), cross-sensor (4\%) as well as cross-dataset scenarios (29\%) compared to state-of-the-art on public domain LivDet datasets.
 
\section*{Acknowledgment}
This research is based upon work supported in part by the Office of the Director of National Intelligence (ODNI), Intelligence Advanced Research Projects Activity (IARPA), via IARPA R\&D Contract No. $2017-17020200004$. The views and conclusions contained herein are those of the authors and should not be interpreted as necessarily representing the official policies, either expressed or implied, of ODNI, IARPA, or the U.S. Government. The U.S. Government is authorized to reproduce and distribute reprints for governmental purposes notwithstanding any copyright annotation therein.

\ifCLASSOPTIONcaptionsoff
  \newpage
\fi



\bibliographystyle{IEEEtran}
\bibliography{fptAntiSpoof.bib}

\begin{thebibliography}{10}
\providecommand{\url}[1]{#1}
\csname url@samestyle\endcsname
\providecommand{\newblock}{\relax}
\providecommand{\bibinfo}[2]{#2}
\providecommand{\BIBentrySTDinterwordspacing}{\spaceskip=0pt\relax}
\providecommand{\BIBentryALTinterwordstretchfactor}{4}
\providecommand{\BIBentryALTinterwordspacing}{\spaceskip=\fontdimen2\font plus
\BIBentryALTinterwordstretchfactor\fontdimen3\font minus
  \fontdimen4\font\relax}
\providecommand{\BIBforeignlanguage}[2]{{%
\expandafter\ifx\csname l@#1\endcsname\relax
\typeout{** WARNING: IEEEtran.bst: No hyphenation pattern has been}%
\typeout{** loaded for the language `#1'. Using the pattern for}%
\typeout{** the default language instead.}%
\else
\language=\csname l@#1\endcsname
\fi
#2}}
\providecommand{\BIBdecl}{\relax}
\BIBdecl

\bibitem{chugh2017fingerprint}
T.~Chugh, K.~Cao, and A.~K. Jain, ``{Fingerprint Spoof Detection using
  Minutiae-based Local Patches},'' in \emph{IEEE International Joint Conference
  on Biometrics (IJCB)}, 2017.

\bibitem{marcel2014handbook}
S.~Marcel, M.~S. Nixon, and S.~Z. Li, \emph{Handbook of Biometric
  Anti-Spoofing}.\hskip 1em plus 0.5em minus 0.4em\relax Springer, 2014.

\bibitem{IARPAProject}
{ODNI, IARPA}, ``{IARPA-BAA-16-04 (Thor)},''
  https://www.iarpa.gov/index.php/research-programs/odin/odin-baa, 2016.

\bibitem{isopad}
{International Standards Organization}, ``{ISO/IEC 30107-1:2016, Information
  Technology---Biometric Presentation Attack Detection---Part 1: Framework},''
  https://www.iso.org/standard/53227.html, 2016.

\bibitem{matsumoto2002impact}
T.~Matsumoto, H.~Matsumoto, K.~Yamada, and S.~Hoshino, ``Impact of artificial
  gummy fingers on fingerprint systems,'' in \emph{Proc. SPIE}, vol. 4677,
  2012, pp. 275--289.

\bibitem{cao2016hacking}
K.~Cao and A.~K. Jain, ``{Hacking mobile phones using 2D Printed
  Fingerprints},'' \emph{MSU Tech. report, MSU-CSE-16-2}, 2016.

\bibitem{arora2016design}
S.~S. Arora, K.~Cao, A.~K. Jain, and N.~G. Paulter, ``{Design and Fabrication
  of 3D Fingerprint Targets},'' \emph{IEEE TIFS}, vol.~11, no.~10, pp.
  2284--2297, 2016.

\bibitem{arora2017goldfingers}
S.~S. Arora, A.~K. Jain, and N.~G. Paulter, ``{Gold Fingers: 3D Targets for
  Evaluating Capacitive Readers},'' \emph{IEEE TIFS}, vol.~12, no.~9, pp.
  2067--2077, 2017.

\bibitem{yoon2012altered}
S.~Yoon, J.~Feng, and A.~K. Jain, ``Altered fingerprints: Analysis and
  detection,'' \emph{IEEE TPAMI}, vol.~34, no.~3, pp. 451--464, 2012.

\bibitem{marasco2015survey}
E.~Marasco and A.~Ross, ``A survey on antispoofing schemes for fingerprint
  recognition systems,'' \emph{ACM Computing Surveys}, vol.~47, no.~2, p.~28,
  2015.

\bibitem{biggio2012security}
B.~Biggio, Z.~Akhtar, G.~Fumera, G.~L. Marcialis, and F.~Roli, ``Security
  evaluation of biometric authentication systems under real spoofing attacks,''
  \emph{IET Biometrics}, vol.~1, no.~1, pp. 11--24, 2012.

\bibitem{yambay2012livdet}
D.~Yambay, L.~Ghiani, P.~Denti, G.~L. Marcialis, F.~Roli, and S.~Schuckers,
  ``{LivDet 2011-Fingerprint liveness detection competition 2011},'' in
  \emph{Proc. 5th IAPR ICB}, 2012, pp. 208--215.

\bibitem{mura2015livdet}
V.~Mura, L.~Ghiani, G.~L. Marcialis, F.~Roli, D.~A. Yambay, and S.~A.
  Schuckers, ``{LivDet 2015 - Fingerprint liveness detection competition
  2015},'' in \emph{Proc. IEEE 7th Intl. Conf. BTAS}, 2015, pp. 1--6.

\bibitem{lapsley1998anti}
P.~D. Lapsley, J.~A. Lee, D.~F. Pare~Jr, and N.~Hoffman, ``Anti-fraud biometric
  scanner that accurately detects blood flow,'' {US Patent 5,737,439, 1998}.

\bibitem{antonelli2006fake}
A.~Antonelli, R.~Cappelli, D.~Maio, and D.~Maltoni, ``Fake finger detection by
  skin distortion analysis,'' \emph{IEEE TIFS}, vol.~1, no.~3, pp. 360--373,
  2006.

\bibitem{baldisserra2006fake}
D.~Baldisserra, A.~Franco, D.~Maio, and D.~Maltoni, ``Fake fingerprint
  detection by odor analysis,'' in \emph{Proc. ICB}.\hskip 1em plus 0.5em minus
  0.4em\relax Springer, 2006, pp. 265--272.

\bibitem{rowe2006multispectral}
R.~K. Rowe and D.~P. Sidlauskas, ``Multispectral biometric sensor,'' Dec.~12
  2006, uS Patent 7,147,153.

\bibitem{hogan2015multiple}
J.~N. Hogan, ``Multiple reference oct system,'' Aug.~25 2015, uS Patent
  9,113,782.

\bibitem{engelsma2017raspireader}
J.~J. Engelsma, K.~Cao, and A.~K. Jain, ``Raspireader: An open source
  fingerprint reader facilitating spoof detection,'' \emph{arXiv preprint
  arXiv:1708.07887}, 2017.

\bibitem{marcialis2010analysis}
G.~L. Marcialis, F.~Roli, and A.~Tidu, ``Analysis of fingerprint pores for
  vitality detection,'' in \emph{Proc. 20th ICPR}, 2010, pp. 1289--1292.

\bibitem{marasco2012combining}
E.~Marasco and C.~Sansone, ``Combining perspiration-and morphology-based static
  features for fingerprint liveness detection,'' \emph{Pattrn. Reco. Letters},
  vol.~33, no.~9, pp. 1148--1156, 2012.

\bibitem{ghiani2012fingerprint}
L.~Ghiani, G.~L. Marcialis, and F.~Roli, ``Fingerprint liveness detection by
  local phase quantization,'' in \emph{Proc. 21st ICPR}, 2012, pp. 537--540.

\bibitem{ghiani2013fingerprint}
L.~Ghiani, A.~Hadid, G.~L. Marcialis, and F.~Roli, ``{Fingerprint liveness
  detection using Binarized Statistical Image Features},'' in \emph{Proc. IEEE
  6th Int. Conf. BTAS}, 2013, pp. 1--6.

\bibitem{gragnaniello2013fingerprint}
D.~Gragnaniello, G.~Poggi, C.~Sansone, and L.~Verdoliva, ``{Fingerprint
  liveness detection based on Weber Local Image Descriptor},'' in \emph{Proc.
  IEEE Workshop on Biometric Meas. Syst. Secur. Med. Appl.}, 2013, pp. 46--50.

\bibitem{gragnaniello2015local}
------, ``Local contrast phase descriptor for fingerprint liveness detection,''
  \emph{Pattern Recognition}, vol.~48, no.~4, pp. 1050--1058, 2015.

\bibitem{menotti2015deep}
D.~Menotti, G.~Chiachia, A.~Pinto, W.~R. Schwartz, H.~Pedrini, A.~X. Falcao,
  and A.~Rocha, ``Deep representations for iris, face, and fingerprint spoofing
  detection,'' \emph{IEEE TIFS}, vol.~10, no.~4, pp. 864--879, 2015.

\bibitem{nogueira2016fingerprint}
R.~F. Nogueira, R.~de~Alencar~Lotufo, and R.~C. Machado, ``{Fingerprint
  Liveness Detection Using Convolutional Neural Networks},'' \emph{IEEE TIFS},
  vol.~11, no.~6, pp. 1206--1213, 2016.

\bibitem{pala2017deep}
F.~Pala and B.~Bhanu, ``{Deep Triplet Embedding Representations for Liveness
  Detection},'' in \emph{Deep Learning for Biometrics. Advances in Computer
  Vision and Pattern Recognition.}\hskip 1em plus 0.5em minus 0.4em\relax
  Springer, 2017, pp. 287--307.

\bibitem{gragnaniello2015investigation}
D.~Gragnaniello, G.~Poggi, C.~Sansone, and L.~Verdoliva, ``An investigation of
  local descriptors for biometric spoofing detection,'' \emph{IEEE TIFS},
  vol.~10, no.~4, pp. 849--863, 2015.

\bibitem{marasco2011robustness}
E.~Marasco and C.~Sansone, ``{On the Robustness of Fingerprint Liveness
  Detection Algorithms against New Materials used for Spoofing},'' in
  \emph{Proc. Intl. Conf. Bio-Insp. Syst. Sign. Process.}, 2011, pp. 553--558.

\bibitem{rattani2015open}
A.~Rattani, W.~J. Scheirer, and A.~Ross, ``Open set fingerprint spoof detection
  across novel fabrication materials,'' \emph{IEEE TIFS}, vol.~10, no.~11, pp.
  2447--2460, 2015.

\bibitem{ding2016ensemble}
Y.~Ding and A.~Ross, ``{An ensemble of one-class SVMs for fingerprint spoof
  detection across different fabrication materials},'' in \emph{Proc. IEEE
  WIFS}, 2016, pp. 1--6.

\bibitem{ghiani2017review}
L.~Ghiani, D.~A. Yambay, V.~Mura, G.~L. Marcialis, F.~Roli, and S.~A.
  Schuckers, ``{Review of the Fingerprint Liveness Detection (LivDet)
  competition series: 2009 to 2015},'' \emph{Image and Vision Computing},
  vol.~58, pp. 110--128, 2017.

\bibitem{russakovsky2015imagenet}
O.~Russakovsky, J.~Deng, H.~Su, J.~Krause, S.~Satheesh, S.~Ma, Z.~Huang,
  A.~Karpathy, A.~Khosla, M.~Bernstein \emph{et~al.}, ``Imagenet large scale
  visual recognition challenge,'' \emph{Proc. IJCV}, vol. 115, no.~3, pp.
  211--252, 2015.

\bibitem{simonyan2014very}
K.~Simonyan and A.~Zisserman, ``Very deep convolutional networks for
  large-scale image recognition,'' \emph{arXiv preprint arXiv:1409.1556}, 2014.

\bibitem{krizhevsky2012imagenet}
A.~Krizhevsky, I.~Sutskever, and G.~E. Hinton, ``{ImageNet classification with
  deep convolutional neural networks},'' in \emph{Proc. Adv. NIPS}, 2012, pp.
  1097--1105.

\bibitem{howard2017mobilenets}
A.~G. Howard, M.~Zhu, B.~Chen, D.~Kalenichenko, W.~Wang, T.~Weyand,
  M.~Andreetto, and H.~Adam, ``Mobilenets: Efficient convolutional neural
  networks for mobile vision applications,'' \emph{arXiv preprint
  arXiv:1704.04861}, 2017.

\bibitem{szegedy2016rethinking}
C.~Szegedy, V.~Vanhoucke, S.~Ioffe, J.~Shlens, and Z.~Wojna, ``{Rethinking the
  Inception Architecture for Computer Vision},'' in \emph{Proc. IEEE CVPR},
  2016, pp. 2818--2826.

\bibitem{ghiani2013livdet}
L.~Ghiani, D.~Yambay, V.~Mura, S.~Tocco, G.~L. Marcialis, F.~Roli, and
  S.~Schuckcrs, ``{LivDet 2013 Fingerprint Liveness Detection Competition
  2013},'' in \emph{Proc. ICB}, 2013, pp. 1--6.

\bibitem{cao2011fingerprint}
K.~Cao, E.~Liu, L.~Pang, J.~Liang, and J.~Tian, ``Fingerprint matching by
  incorporating minutiae discriminability,'' in \emph{Proc. IJCB}, 2011, pp.
  1--6.

\bibitem{chollet2016xception}
F.~Chollet, ``Xception: Deep learning with depthwise separable convolutions,''
  \emph{arXiv preprint arXiv:1610.02357}, 2016.

\bibitem{roy2017masterprint}
A.~Roy, N.~Memon, and A.~Ross, ``Masterprint: Exploring the vulnerability of
  partial fingerprint-based authentication systems,'' \emph{IEEE Transactions
  on Information Forensics and Security}, vol.~12, no.~9, pp. 2013--2025, 2017.

\end{thebibliography}

\begin{IEEEbiography}[{\includegraphics[width=1in,height=1.25in,clip,keepaspectratio]{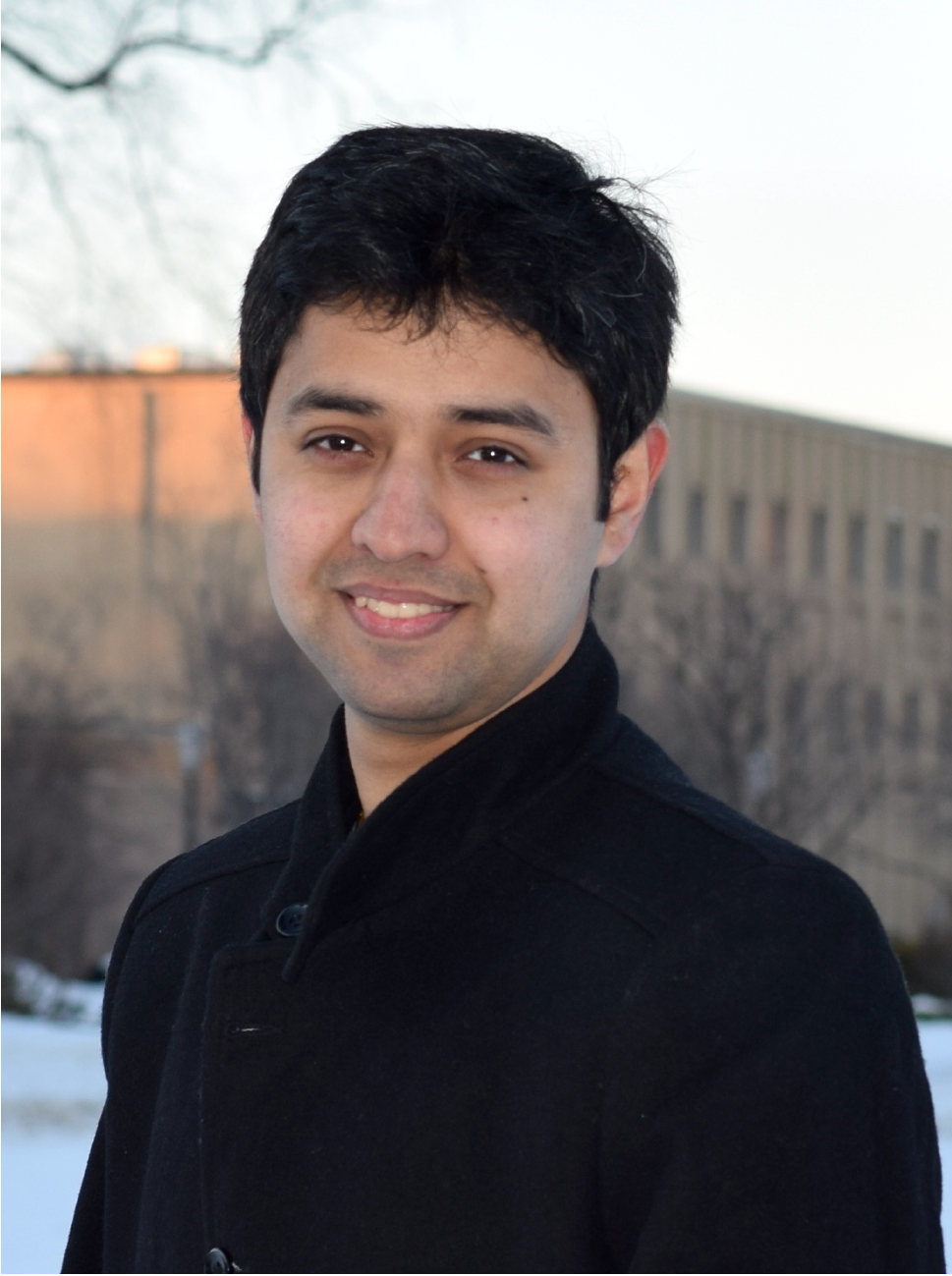}}]{Tarang Chugh}
received the B. Tech. (Hons.) degree in Computer Science and Engineering from the Indraprastha Institute of Information Technology, Delhi (IIIT-D) in 2013. He was affiliated with IBM Research Lab, New Delhi, India as a research engineer during 2013-2015. He is currently a doctoral student in the Department of Computer Science and Engineering at Michigan State University. His research interests include biometrics, pattern recognition, and machine learning.
\end{IEEEbiography}

\begin{IEEEbiography}[{\includegraphics[width=1in,height=1.25in,clip,keepaspectratio]{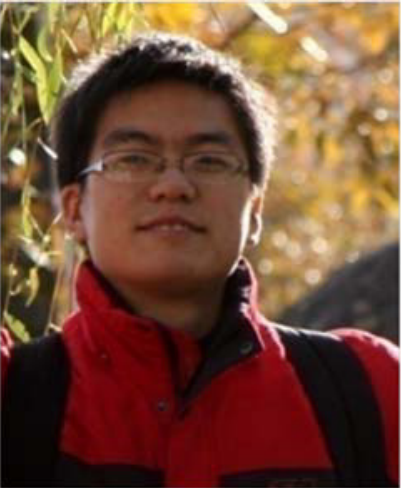}}]{Kai Cao}
received the Ph.D. degree from the Key Laboratory of Complex Systems and Intelligence Science, Institute of Automation, Chinese Academy of Sciences, Beijing, China, in 2010. He is currently a Post Doctoral Fellow in the Department of Computer Science \& Engineering, Michigan State University. He was affiliated with Xidian University as an Associate Professor. His research interests include biometric recognition, image processing and machine learning.
\end{IEEEbiography}

\begin{IEEEbiography}[{\includegraphics[width=1in,height=1.25in,clip,keepaspectratio]{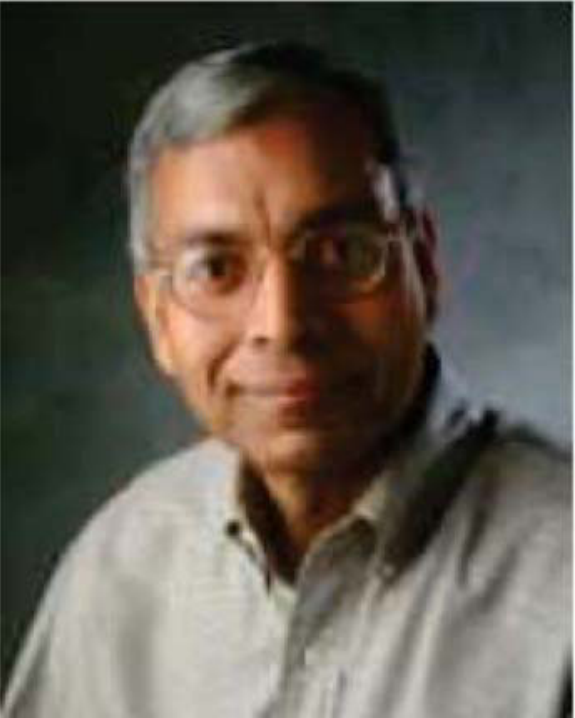}}]{Anil K. Jain}
is a University distinguished professor in the Department of Computer Science and Engineering at Michigan State University. His research interests include pattern recognition and biometric authentication. He served as the editor-in-chief of the IEEE Transactions on Pattern Analysis and Machine Intelligence and was a member of the United States Defense Science Board. He has received Fulbright, Guggenheim, Alexander von Humboldt, and IAPR King Sun Fu awards. He is a member of the National Academy of Engineering and foreign fellow of the Indian National Academy of Engineering.
\end{IEEEbiography}

\end{document}